\documentclass[review]{elsarticle}
\usepackage{subfigure}
\usepackage{amssymb}
\usepackage{amsthm}
\usepackage{amsmath}
\usepackage{amsfonts}
\usepackage{graphicx}
\usepackage{caption}
\usepackage{amsmath,graphicx,subfigure,threeparttable,booktabs,float,stfloats,multirow}
\usepackage{xcolor}
\usepackage{booktabs}
\usepackage{multirow}
\usepackage{diagbox}
\usepackage{bm}
\usepackage{url}
\usepackage{algorithm}
\usepackage{algorithmic}
\usepackage{graphics}
\usepackage{indentfirst}
\usepackage{multirow}

\begin{document}

\begin{frontmatter}

\title{Compressed Sensing MRI via a Multi-scale Dilated Residual Convolution Network}
\author[1]{Yuxiang Dai}
\author[1]{Peixian Zhuang*}
\ead{zhuangpeixian0624@163.com}

\address{$^1$Jiangsu Key Laboratory of Meteorological Observation and Information Processing,\\
$^1$Jiangsu Technology and Engineering Center of Meteorological Sensor Network, \\
$^1$School of Electronic and Information Engineering,\\
$^1$Nanjing University of Information Science and Technology, Nanjing 210044, China}



\begin{abstract}
Magnetic resonance imaging (MRI) reconstruction is an active inverse problem which can be addressed by conventional compressed sensing (CS) MRI algorithms that exploit the sparse nature of MRI in an iterative optimization-based manner. However, two main drawbacks of iterative optimization-based CSMRI methods are time-consuming and are limited in model capacity. Meanwhile, one main challenge for recent deep learning-based CSMRI is the trade-off between model performance and network size. To address the above issues, we develop a new multi-scale dilated network for MRI reconstruction with high speed and outstanding performance. Comparing to convolutional kernels with same receptive fields, dilated convolutions reduce network parameters with smaller kernels and expand receptive fields of kernels to obtain almost same information. To maintain the abundance of features, we present global and local residual learnings to extract more image edges and details. Then we utilize concatenation layers to fuse multi-scale features and residual learnings for better reconstruction. Compared with several non-deep and deep learning CSMRI algorithms, the proposed method yields better reconstruction accuracy and noticeable visual improvements. In addition, we perform the noisy setting to verify the model stability, and then extend the proposed model on a MRI super-resolution task.
\end{abstract}


\end{frontmatter}

\section{INTRODUCTION}
MRI is a widely used imaging technology for visualizing the structure and functioning of the body with the advantages of non-radiation and non-ionizing nature. However, the slow imaging speed of MR poses a limitation on its widespread application. Recently, the CS theory [1] is introduced to reduce the MR scan time, and CSMRI can reconstruct a high resolution image from randomly sampled \emph{k}-space data. The CSMRI problem can be formulated as the optimization
\begin{small}
\begin{equation}
\begin{array}{l}
\begin{split}
&\hat{x}=\mathop{\arg min}_{\emph{x}}\frac{1}{2}{\|{F_u}\emph{x}-\emph{y}\|}_2^2+{\Sigma_i}{\alpha_i}{R_i(x)}
\end{split}
\end{array}
\end{equation}
\end{small}
Where \emph{x} is denoted as the MRI to be reconstructed, \emph{y} are the \emph{k}-space data, and ${F_u}$ represents the under-sampled Fourier encoding matrix. The first term ${\|{F_u}\emph{x}-\emph{y}\|}_2^2$ indicates data fidelity that can ensure the consistence between the Fourier coefficients of the reconstructed image and measured data. The second term ${R_i}$ is an analytical, sparsifying transform term, and ${\alpha_i}$ is a factor for balancing data fidelity and transform terms. MR images can be generated by inverse Fourier transform of the sampled \emph{k}-space data, which are the Fourier coefficient of an object. However, aliasing artifacts (noise-like) are produced by the incoherence of under-sampled \emph{k}-space in transform domain, as shown in Fig. 1.
\begin{figure*}[!t]
\graphicspath{{figure15/}}
\centering
\subfigure[]
{\includegraphics[height=1in,width=1in,angle=0]{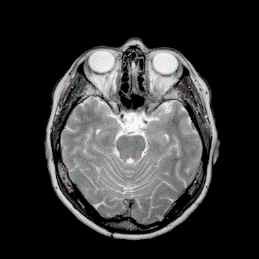}}
\ \ \ \ \subfigure[]
{\includegraphics[height=1in,width=1in,angle=0]{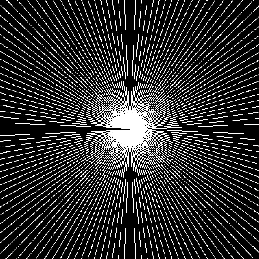}}
\ \ \ \ \subfigure[]
{\includegraphics[height=1in,width=1in,angle=0]{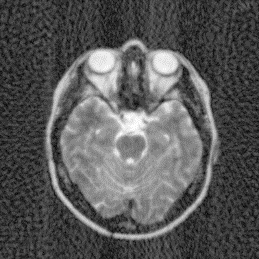}}
\ \ \ \ \subfigure[]
{\includegraphics[height=1in,width=1in,angle=0]{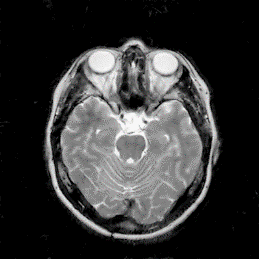}}
\caption{The zero-filled reconstruction. (a) is a full-sampled MRI, (b) is a 20\% radial sampling mask, (c) is the zero-filled reconstruction under (b), and (d) is the reconstruction using our method. Note that aliasing artifacts are clearly seen in the zero-filled reconstruction (c), which impair diagnostic information. However, our algorithm can remove these unpleasant artifacts (d).}
\end{figure*}

To address this problem, a large number of CSMRI algorithms have been developed, and these methods tend to fall into two main categories:

The first category of CSMRI algorithms are iterative optimization-based CSMRI, in which the sparsity is enforced in specific transform domain or underlying latent representation of images, and then an alternating iterative optimization scheme is adopted to CSMRI reconstruction [2]-[11]. A pioneering work of CSMRI is Sparse MRI [2], which exploits an off-the-shelf basis to capture a specific feature (wavelets recover point-like features, contourlets capture curve-like features). A hybrid TV regularizer combined with a $L_0$-regularized tree-structured sparsity constraint [3] is introduced to overcome model-dependent deficiency and represent the measure of sparseness in wavelet domain. However, fixed bases fail to sparsely represent complicated MR images with underlying image edges and textures. To address this issue, several dictionary learning models (DLMRI [4], BPFA [5] and FDLCP [6]) and different wavelet regularizations based on geometric information (PBDW [7] and PBDW with pFISTA [8]) are exploited. For instance, a fast orthogonal dictionary learning method (FDLCP) is introduced to provide adaptive sparse representation of images, in which image is divided into classified patches according to the same geometrical direction and dictionary is trained within each class for enhanced sparsity. And patch-based directional wavelets model (PBDW) is proposed to promote MRI reconstruction, and patch geometric direction is trained from the reconstructed image using conventional CSMRI methods. But these dictionary learning or wavelet regularizations algorithms are required that parameters such as dictionary size and patch sparsity are preset. A Bayesian non-parametric dictionary learning model (BPTV) [9] applies the beta process to learn the sparse representation necessary for CSMRI, in which beta process is an effective prior for non-parametric learning of dictionary parameters such as dictionary size and patch sparsity. In addition, some methods are performed to obtain the information from the MRI of interest. A method (PANO) [10] exploits nonlocal similarity of image patches by establishing a patch-based nonlocal operator, which effectively produces sparse vectors by operating on grouped similar patches of the image. Another MRI reconstruction algorithm can promote structures and suppress artifacts with an edge-preserving filtering prior [11], in which a gradient domain guided image filtering (GFF) is embedded. However, conventional CSMRI methods are limited in model capacity to recover diverse image structures, and require a lot of iterative operations which is time-consuming and fails in real-time reconstructions.

The deep learning-based CSMRI [12]-[16] can learn a nonlinear mapping from the zero-filled MRI to the fully-sampled MRI. In addition, better MR images can be reconstructed by exploiting existing training mode with no additional iterations, which can achieve real-time execution compared with iterative optimization-based CSMRI. For the purpose of accelerating MR imaging, an off-line convolutional neural network (CNN) [12] is applied for CSMRI by learning an end-to-end mapping between zero-filled and fully-sampled MR images for the first time. After that, a deep cascade of CNNs [13] combines convolution and data sharing approaches to identify spatio-temporal correlations in MR images, which can boost data acquisition. In order to accelerate MR acquisition process with performance guarantee, U-net with deep residual learning [14] is proposed to formulate a CS problem as a residual regression problem where aliasing artifacts from under-sampled data are simpler than those of images in structure. ADMM-Net [15] uses alternating direction method of multiplies (ADMM) to derive and define the data flow, which can optimize a general CSMRI model to reconstruct MR images from a small number of under-sampled data in \emph{k}-space. Moreover, in the Bayesian deep learning model [16], the MC-dropout and heteroscedastic loss are applied to the reconstruction networks to model epistemic and aleatoric uncertainty which can achieve competitive performance. Although the above-mentioned deep learning algorithms can accelerate MR acquisition process with performance guarantee, they are composed of complex network with more parameters.

To address these limitations, we develop a novel multi-scale dilated network (MDN) for MRI reconstruction. The contributions of our paper are summarized as follows:
\begin{itemize}
 \item We develop a dilated network to expand the receptive field of convolutional kernel for reducing network parameters without the loss of resolution, which can obtain multi-scale information. When compared with the larger kernels with same receptive field, the dilated network can increase reconstruction accuracy and accelerate training speed.
 \item Considering the structures and details synthetically, we adopt global residual learning to make up the overall structural features missed during extracting process, and employ local residual learnings to extract more abundant features to preserve better edges and details.
 \item We exploit concatenation layers to fuse multi-scale features, which can make full use of the abundance of features to maintain image details for better reconstruction results.
 \item We perform numerous experiments to demonstrate the better capability of the proposed model with three sampling masks and a variety of sampling rates for each mask. In addition, the proposed model can be applied into MRI noisy setting and super-resolution tasks, which demonstrate the effectiveness of the proposed model.
\end{itemize}

\section{RELATED WORK}
In this section, we review the related components of deep learning that are used in the proposed network for MRI reconstruction.

\textbf{\emph{Residual learning}}: As the number of network layer increased, the expression of the overall model is enhanced, which results in poor training accuracy. The deep residual network [17] introduces an equal fast connection to solve the problem of gradient disappearance. The basic block of residual is to use a shortcut during two contiguous convolutional layers. Residual learning has achieved impressive performance on image low-level tasks, such as reconstruction [18]-[21], super-resolution [22]-[24], denoising [25]-[27], deraining [28]-[30], etc.

\textbf{\emph{Dilated convolution}}: Dilated convolution (conv) has been proposed for intensive prediction tasks [31]-[33]. Compared with the ordinary convolution, dilated convolution has a dilated rate parameter called dilated factor (\emph{DF}), which is mainly used to indicate the dilatation size. Dilated convolution has a larger receptive field, while keeping the number of kernel parameters constant with the same size as the ordinary convolution. The feature map size of the output can be stayed the same by dilated convolution.

\textbf{\emph{Concatenation}}: The concatenation (concat) layer [34]-[36] is used to splice two or more feature maps in the channel or number dimension without operating the residual layer, which can fuse single-scale or multi-scale features. For instance, when conv\_1 and conv\_2 are spliced on the channel dimension (${k_1}$,${k_2}$), the other dimensions (\emph{${N_i}$}, \emph{H}, and \emph{W}) must be consistent, where \emph{${N_i}$} is the number of image patches, \emph{H} and \emph{W} represent the height and width of output matrix respectively. The operation at this time is channel ${k_1}$ plus channel ${k_2}$ and the output of the concat layer can be expressed as: \emph{${N_i}$}$\times$(${k_1}$+${k_2}$)$\times$\emph{H}$\times$\emph{W}. The concat layer is generally used to employ the semantic information of multi-scale feature maps to achieve better performance.

\section{METHOD}

\noindent
\textbf{A.Problem formulation}

Different from traditional CSMRI problem, deep learning-based algorithms can generate a pre-trained model which can be directly transferred in MR imaging. And the deep learning-based algorithms require to train huge measurements for the model with optimal performance, which can be seen in Fig. 2. The deep learning-based CSMRI problem can be formulated as:
\begin{small}
\begin{equation}
\begin{array}{l}
\begin{split}
&{\hat{x}}=\mathop{\arg min}_{\emph{x}}\frac{1}{2}{\|{F_u}\emph{x}-\emph{ y}\|}_2^2+\xi{\|x-{f_{cnn}({x_u}|{\hat{\theta}})}\|}_2^2
\end{split}
\end{array}
\end{equation}
\end{small}
Where ${f_{cnn}}$ is the forward propagation of the CNN with the parameter $\theta$ that contains millions of network weights, and $\xi$ is a regularization parameter. ${x_u}$ represents the zero-filled images as shown in Fig. 2, and ${\hat{\theta}}$ are the optimal parameters of trained CNN.

\begin{figure*}[!t]
\graphicspath{{figure15/}}
\centering
\subfigure[]
{\includegraphics[height=1.45in,width=5in,angle=0]{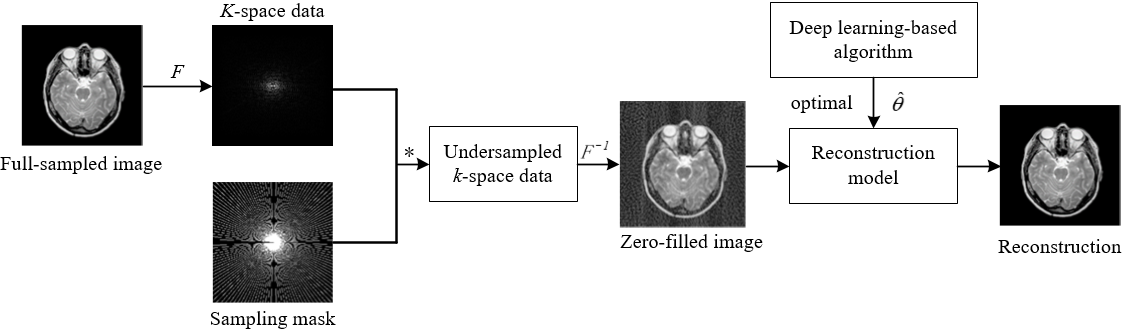}}
\caption{The flowchart of the MR imaging based on deep learning approaches. \emph{F} is Fourier transform, * is the process of under-sampling on \emph{k}-space data and $F^{-1}$ represents inverse Fourier transform.}
\end{figure*}

\noindent
\textbf{B.Proposed block}

A multi-scale dilated network (MDN) block consists of dilated convolution with rectified linear units (relu) [37], residual learning and multi-scale concatenation. Fig 4 shows the overall framework of the proposed MDN architecture. Then we present the compositions of the proposed block in detail.

\textbf{\emph{Dilated convolution}}: As shown in Fig. 4, there are 7 convolutional layers in a MDN-block, which consists of one normal convolution, three 2-dilated convolutions and three 3-dialted convolutions. As we all know, a convolutional layer can extract \emph{n} layers of features when the \emph{${N_f}$} is set to \emph{n}, in which \emph{${N_f}$} is the number of filters (convolutional kernels) in the convolutional layers. It is well-known that more features are extracted as the number of kernels becomes larger, and the effect of network training increases accordingly. For reducing the parameters to lower the computational complexity, we choose proper \emph{DF} and \emph{${N_f}$} for convolutional layers. In Fig. 4, \emph{DF} is set to 3 when \emph{${N_f}$} is 32, conversely, \emph{DF} is set to 2 when \emph{${N_f}$} is 64.

We increase the receptive field of the convolutional kernel to expand the receiving domain of image information. Since the number of convolution kernels is limited, appropriate \emph{DF} should be chosen to match the proposed network and data sets. Compared with the original convolutional layer (the size of kernel is increased), the dilated convolution achieves comparable performance with less parameters demonstrated in experiments. The key is to obtain a better tradeoff among the number, size and dilated factor of convolutional kernels.

\textbf{\emph{Local and global residuals}}: The global and local residual learnings are integrated to maintain the abundance of feature maps for better reconstruction. Global residual learning (GRL) tries to obtain initial information, while local residual learnings (LRLs) are utilized to further improve the information flow.
\begin{figure*}[!t]
\graphicspath{{figure15/}}
\centering
\subfigure[GRL]
{\includegraphics[height=0.9in,width=5in,angle=0]{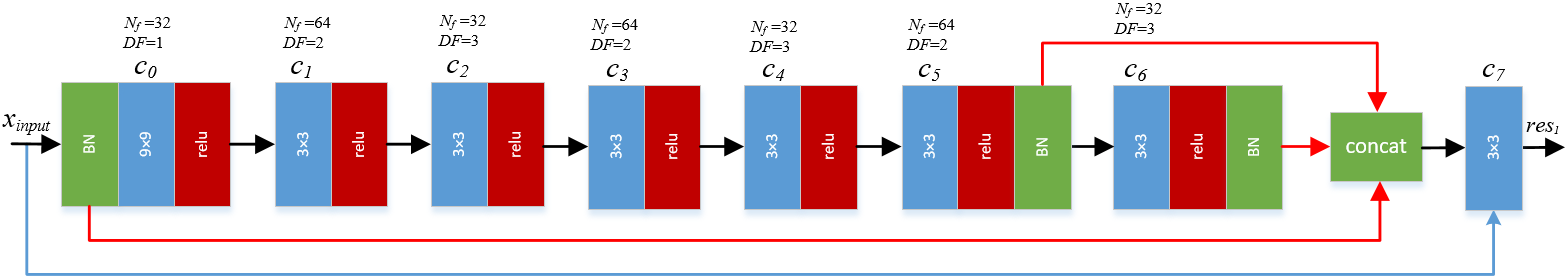}}
\subfigure[LRLs]
{\includegraphics[height=1in,width=5in,angle=0]{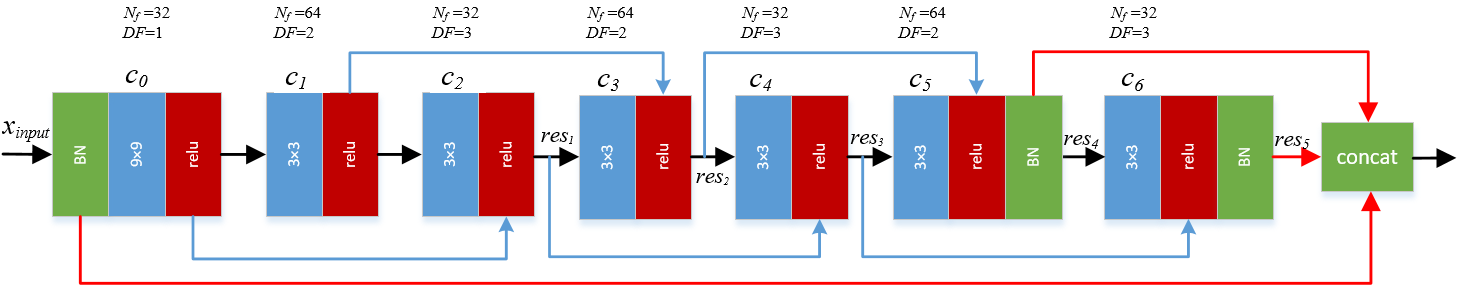}}
\caption{ GRL and LRLs based on dilated network. (a) is the GRL, which takes advantage of initial information. (b) is the LRLs, which further improves the information flow. Then MDN integrates both of them.}
\end{figure*}
Fig. 3(a) shows that the GRL concatenates a series of convolution layers, and finally connects the input to the output for preventing the loss of features. Fig. 3(b) presents the LRLs where there are five local residuals in a block, which do not burden the network complexity. Fig. 4 is the proposed residual network. We combine GRL and LRLs to ensure adequate features, slightly increase the network complexity, and achieve better reconstruction results than the above two residual learnings.\\
GRL:
\begin{small}
\begin{equation}
\begin{array}{l}
\begin{split}
{c_n}=f({c_{n-1}}), res_1=I_{input}+ {c_7}
\end{split}
\end{array}
\end{equation}
\end{small}
LRLs:
\begin{small}
\begin{equation}
\begin{array}{l}
\begin{split}
\\res_1&={c_0}+{c_2}, res_2={c_1}+ {c_3},
\\res_3&=res_1+{c_4}, res_4=res_2+ {c_5},
\\res_5&=res_3+{c_6}.
\end{split}
\end{array}
\end{equation}
\end{small}
Proposed Residual:
\begin{small}
\begin{equation}
\begin{array}{l}
\begin{split}
\\res_1&={c_0}+{c_2}, res_2={c_1}+ {c_3},
\\res_3&=res_1+{c_4}, res_4=res_2+ {c_5},
\\res_5&=res_3+{c_6}, res_6=I_{input}+ {c_7}.
\end{split}
\end{array}
\end{equation}
\end{small}

Where \emph{f}($\cdot$) represents the operation of convolutional layer and activation function, ${c_n}$ denotes feature maps of the \emph{n}-th convolutional layer, $res_n$ is the \emph{n}-th residual sum, and $I_{input}$ represents the input images. The proposed residual learnings have a better effect on MRI reconstruction without burdening the network complexity. We utilize both GRL and LRLs in the network to prevent the loss of valid features.

\textbf{\emph{Multi-scale Concatenation}}:
The computational complexity of residual block with dilated convolutions shows a very high growth trend, especially in the case of huge data sets. To solve the shortcoming, we exploit a multi-scale residual block, in which different numbers of convolutional kernels and residual learnings are integrated to enrich features. At the same time, multi-scale features are stacked so that abundant information can be shared and reused, which contributes to the fusion of local features. In addition, the application of a 3$\times$3 kernel after concat (that is after a block) aims to facilitate the fusion of features and cut down computational complexity, and batch normalization [38] is utilized before the input of concat to improve accuracy and accelerate convergence, which can accelerate MR imaging.

\begin{figure*}[!t]
\graphicspath{{figure15/}}
\centering
{\includegraphics[height=2.5in,width=4.7in,angle=0]{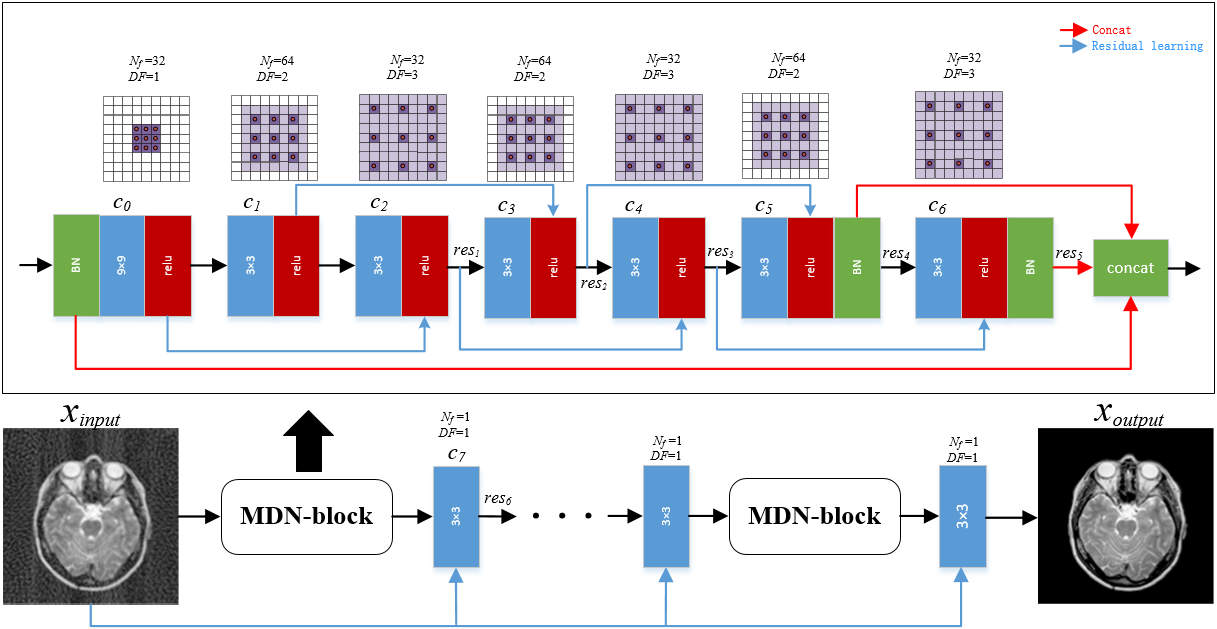}}
\caption{Schema for the proposed MDN for CSMRI. The light purple regions stand for the receptive fields of the dilated convolutional kernels with different \emph{DF} values. The blue lines are the input of residual leanings, and the red ones are the input of a concatenation layer.}
\end{figure*}

\noindent
\textbf{C.Network architecture}

As discussed above, the proposed MDN framework consists of repeated blocks. The residual sum after a block aims to supplement initial information missing in the process of extracting features during the previous blocks (Fig. 4). However, deeper the network with more blocks does not mean that the extracted features are more favorable to reconstruction results. Proper number of blocks should be adopted for CSMRI.

The loss function of the proposed network is:
\begin{small}
\begin{equation}
\begin{array}{l}
\begin{split}
Loss=\frac{1}{2M}\sum_{i=1}^M({\hat{x}_i}-{x_i})^2
\end{split}
\end{array}
\end{equation}
\end{small}
where {$x_i$} denotes full-sampled image, and {$\hat{x}_i$} represents the output of the network; \emph{M} is the number of training images. The proposed network can be implemented using Caffe, Pytorch or Tensorflow.

\section{EXPERIMENT RESULTS}

\begin{figure*}[!t]
\graphicspath{{figure15/}}
\centering
\ \ \ \ \ \ \ \subfigure[]
{\includegraphics[height=1.2in,width=1.2in,angle=0]{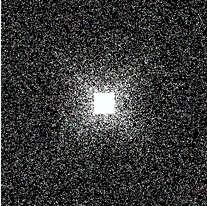}}
\ \ \ \ \ \ \ \subfigure[]
{\includegraphics[height=1.2in,width=1.2in,angle=0]{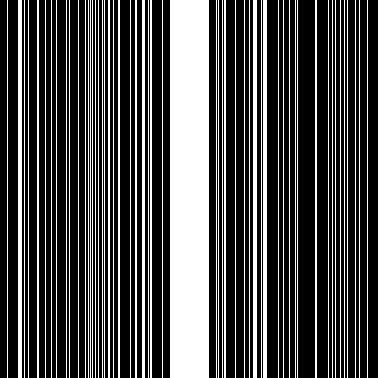}}
\ \ \ \ \ \ \ \subfigure[]
{\includegraphics[height=1.2in,width=1.2in,angle=0]{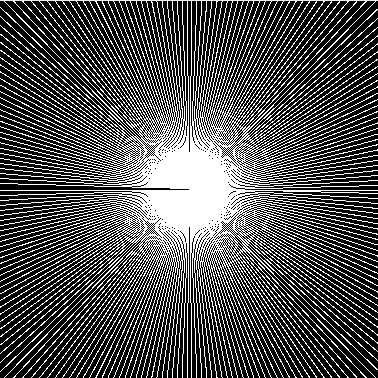}}
\caption{Three sampling masks with specfic sampling rates. (a) 20\% variable density random sampling. (b) 25\% cartesian sampling. (3) 30\% radial sampling.}
\end{figure*}

We provide numerous experiments to demonstrate the effectiveness of the proposed method in MR reconstruction, which compares with several iterative optimization-based and deep learning-based approaches. We employ three sampling masks: variable density sampling [2], cartesian sampling [39] and radial sampling [40], and a variety of sampling rates are set for each mask. An example of each mask is shown in Fig. 5. Then we consider the noisy settings and apply the proposed model into MR super-resolution. In addition, the ablation study on residual learnings is conducted to illustrate the effect of GRL and LRLs, and different initial learning rates are considered in experiments.

\emph{\textbf{Implementation details}} We train and test the network based on the NVIDIA GeForce GTX 1080Ti with 11GB GPU memory. We use Caffe for algorithm in network training and Matlab for preprocessing of data sets. The maximum iterations of the network are 250,000. The main function of the solver in Caffe is to alternately transfer forward and backward conduction to update the weight of neural network, so as to minimize the loss. The optimization we used is ‘Adam’, the base learning rate is set to 0.001, and the weight decay is set to 0.0001, which is weight attenuation term used to prevent over-fitting. The learning rate policy is ‘step’ and the gamma coefficient associated with the learning rate is set to 0.1. The step size indicates the frequency at which we should go to the next ‘training step’, which is set to 50000. The weight of the last gradient update (momentum) is set to 0.9. {The training errors are displayed per 100 iterations and testing errors are displayed per epoch, which can be seen in Fig. 6.}

\begin{figure*}[!t]
\graphicspath{{figure15/}}
\centering
\subfigure[]
{\includegraphics[height=1.5in,width=2in,angle=0]{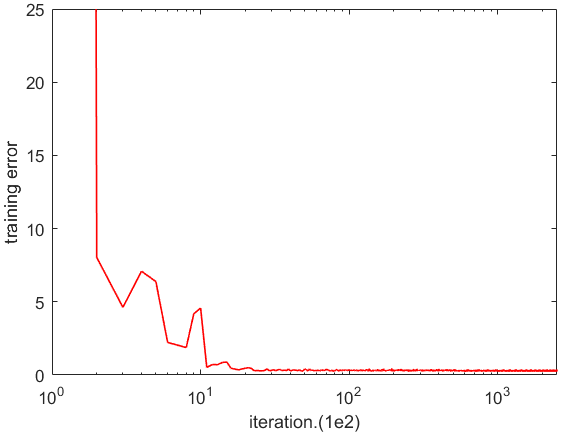}}
\ \ \ \ \subfigure[]
{\includegraphics[height=1.5in,width=2in,angle=0]{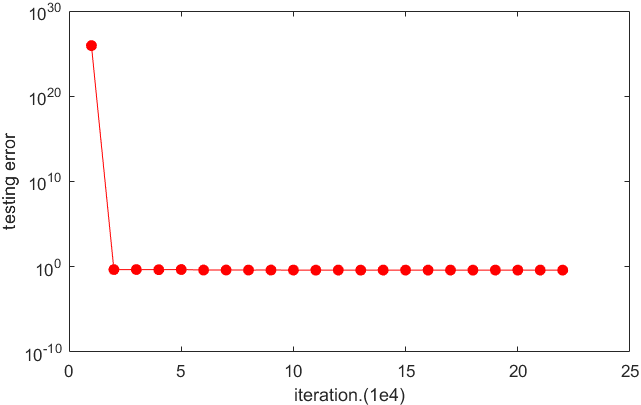}}
\caption{The convergence curves of our network. (a) Training convergence curve. (b) Testing convergence curve.}
\end{figure*}

\textbf{\emph{Data sets}} Our real-valued data sets come from the MRI Multiple Sclerosis Database (MRI MS DB)\footnotemark[1]\footnotetext[1]{http://www.medinfo.cs.ucy.ac.cy/index.php/facilities/32-software/218-datasets}. Among them, we select 450 T2-MR images as a training set. In addition, we expand this training set to 1534 images by rotating these 450 pictures, and we consider 50 high quality T2 images as a test set. Moreover, we choose 800 simulated complex images as a train set, and select 80 simulated complex  images as a test set. All images have the size of 378$\times$378.

\textbf{\emph{Metrics}} We not only evaluate the reconstructed results subjectively, but also use two objective evaluation indicators: peak signal to noise ratio (PSNR) [41] and structural similarity index (SSIM) [42]. The PSNR represents the ratio between the power of the maximum possible image intensity across a volume and the power of distorting noise and other errors, and the SSIM shows the similarity between two images by exploiting the inter-dependencies among nearby pixels. Higher values of PSNR and SSIM demonstrate better reconstruction. Additionally, we employ the standard deviation of PSNR to demonstrate the network stability on  complex-valued data.

\textbf{\emph{Quantitative evaluation}} To evaluate the reconstruction performance, we compare the proposed model with the two iterative optimization-based methods: Sparse MRI [2] and DLMRI [4], and three deep-learning algorithms: Single-scale residual learning (Single-scale) [14], LRLs and U-net [14]. The former two optimizations are provided by the authors' homepage. The latter three deep learning algorithms are reproduced using Caffe. We consider the zero-filled reconstruction results as well. We reproduce Single-scale residual learning in the same environment, which uses a modified deconvolution network with symmetric contracting path. Based on Single-scale residual learning, the U-net utilizes the pooling layer and deconvolution to make full use of multi-scale features. LRLs has been shown in Fig. 3(b).
\subsection{\textbf{Experiments on real-valued MRI with different masks}}
\begin{table*}[t]\tiny
\centering
 \begin{threeparttable}
 \caption{\label{tab:results} PSNR/SSIM of different methods on real-valued brain MRI as a function of sampling percentage. Sampling masks include cartesian sampling, variable density random sampling and radial sampling. The value with red bold font indicates ranking the first place while value with blue font is the second place.}
  \begin{tabular}{cccccccc}
  \toprule
  Mask &Sampling\% &Sparse MRI &DLMRI &Single-scale &LRLs &U-net & MDN \\
  \midrule
   Cartesian &10 &24.50/0.811 &25.22/0.726 &25.46/0.797 &26.14/0.802 &\textcolor[rgb]{0.00,0.07,1.00}{26.15/0.815} &\textcolor[rgb]{1.00,0.00,0.00}{\textbf{26.59/0.840}}\\
             &15 &26.16/0.857 &28.37/0.841 &28.29/\textcolor[rgb]{0.00,0.07,1.00}{0.861} &\textcolor[rgb]{0.00,0.07,1.00}{28.62}/0.860 &28.29/0.850 &\textcolor[rgb]{1.00,0.00,0.00}{\textbf{28.86/0.871}}\\
             &20 &26.98/0.885 &30.68/0.902 &30.53/0.905 &\textcolor[rgb]{0.00,0.00,1.00}{31.05/0.907} &30.80/\textcolor[rgb]{0.00,0.07,1.00}{0.907} &\textcolor[rgb]{1.00,0.00,0.00}{\textbf{31.43/0.930}}\\
             &25 &27.60/0.895 &32.85/0.934 &32.30/0.930 &32.47/0.931 &\textcolor[rgb]{0.00,0.07,1.00}{33.13/0.939} &\textcolor[rgb]{1.00,0.00,0.00}{\textbf{33.25/0.950}}\\
             &30 &28.45/0.892 &34.77/0.955 &34.28/0.954 &34.57/0.954 &\textcolor[rgb]{0.00,0.07,1.00}{34.81/0.958} &\textcolor[rgb]{1.00,0.00,0.00}{\textbf{35.27/0.967}}\\
 \midrule
   Random    &10 &27.38/0.776 &31.27/0.554 &\textcolor[rgb]{0.00,0.07,1.00}{32.07/0.904} &31.51/0.887 &32.01/0.902 &\textcolor[rgb]{1.00,0.00,0.00}{\textbf{32.16/0.913}}\\
             &15 &27.64/0.821 &32.86/0.612 &32.76/0.908 &33.15/0.876 &\textcolor[rgb]{0.00,0.07,1.00}{33.22/0.920} &\textcolor[rgb]{1.00,0.00,0.00}{\textbf{33.82/0.930}}\\
             &20 &30.44/0.888 &34.34/0.675 &33.99/0.924 &34.17/0.919 &\textcolor[rgb]{0.00,0.07,1.00}{34.70/0.942} &\textcolor[rgb]{1.00,0.00,0.00}{\textbf{34.95/0.944}}\\
             &25 &33.44/0.915 &35.75/0.727 &34.97/0.939 &35.02/0.928 &\textcolor[rgb]{0.00,0.07,1.00}{35.77/0.947} &\textcolor[rgb]{1.00,0.00,0.00}{\textbf{35.96/0.948}}\\
             &30 &34.71/\textcolor[rgb]{1.00,0.00,0.00}{\textbf{0.963}} &36.75/0.754 &35.86/0.949 &35.74/0.936 &\textcolor[rgb]{0.00,0.07,1.00}{36.63}/0.954 &\textcolor[rgb]{1.00,0.00,0.00}{\textbf{36.83}}/\textcolor[rgb]{0.00,0.07,1.00}{0.958}\\
 \midrule
   Radial    &10 &23.17/0.668 &27.93/0.405 &28.19/0.815 &28.98/0.844 &\textcolor[rgb]{0.00,0.07,1.00}{29.00/0.849} &\textcolor[rgb]{1.00,0.00,0.00}{\textbf{29.64/0.873}}\\
             &15 &24.68/0.742 &29.77/0.448 &30.26/0.860 &\textcolor[rgb]{0.00,0.07,1.00}{30.96/0.877} &30.67/0.871 &\textcolor[rgb]{1.00,0.00,0.00}{\textbf{31.87/0.905}}\\
             &20 &25.91/0.648 &30.55/0.467 &31.97/0.888 &\textcolor[rgb]{0.00,0.07,1.00}{32.54}/0.888 &32.48/\textcolor[rgb]{0.00,0.07,1.00}{0.889} &\textcolor[rgb]{1.00,0.00,0.00}{\textbf{33.48/0.925}}\\
             &25 &26.14/0.773 &31.02/0.478 &33.21/0.917 &33.75/0.921 &\textcolor[rgb]{0.00,0.07,1.00}{33.84/0.925} &\textcolor[rgb]{1.00,0.00,0.00}{\textbf{34.51/0.938}}\\
             &30 &28.26/0.898 &31.35/0.487 &34.21/\textcolor[rgb]{0.00,0.07,1.00}{0.935} &34.91/0.928 &\textcolor[rgb]{0.00,0.07,1.00}{35.12}/0.932 &\textcolor[rgb]{1.00,0.00,0.00}{\textbf{35.64/0.955}}\\
  \bottomrule
  \end{tabular}
 \end{threeparttable}
\end{table*}
\begin{figure*}[!t]
\graphicspath{{figure15/}}
\centering
\subfigure[Ground truth]
{\includegraphics[height=1in,width=1in,angle=0]{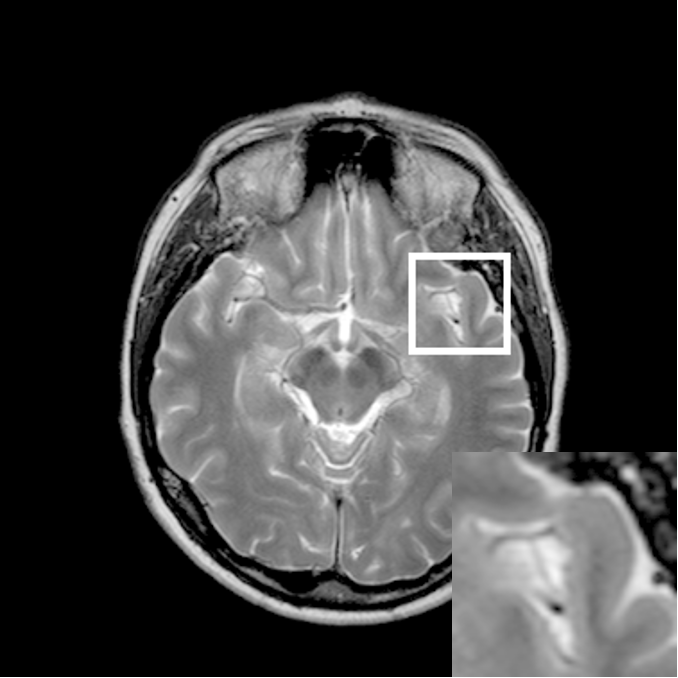}}
\ \ \ \ \ \subfigure[Zero-filling]
{\includegraphics[height=1in,width=1in,angle=0]{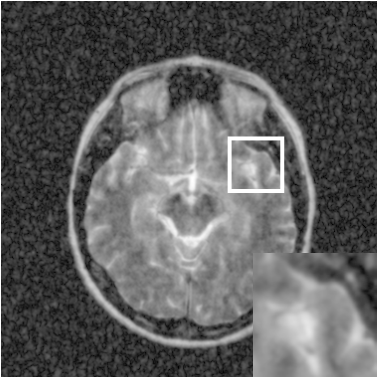}}
\ \ \ \ \ \subfigure[Sparse MRI]
{\includegraphics[height=1in,width=1in,angle=0]{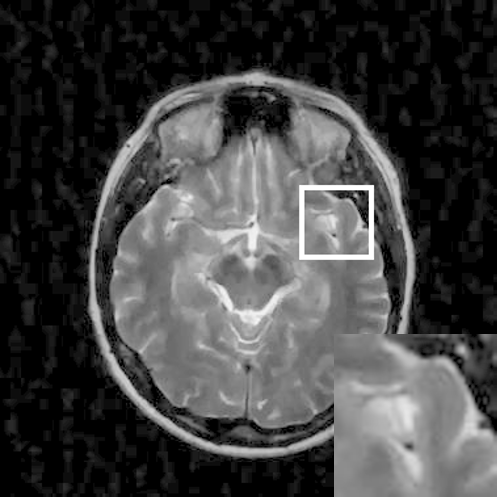}}
\ \ \ \ \ \subfigure[DLMRI]
{\includegraphics[height=1in,width=1in,angle=0]{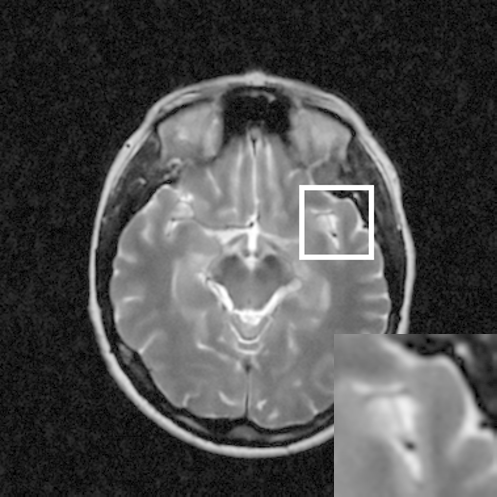}}
\ \ \ \ \ \subfigure[Single-scale]
{\includegraphics[height=1in,width=1in,angle=0]{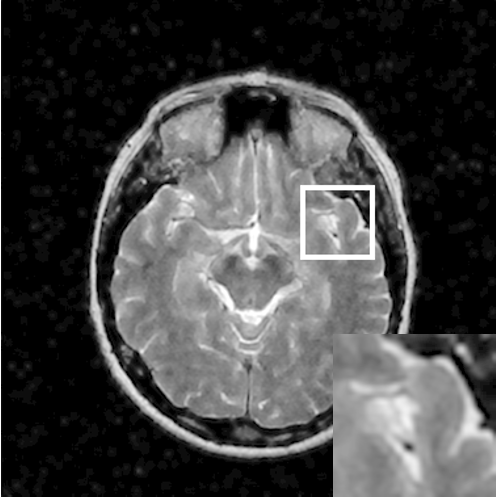}}
\ \ \ \ \ \subfigure[LRLs]
{\includegraphics[height=1in,width=1in,angle=0]{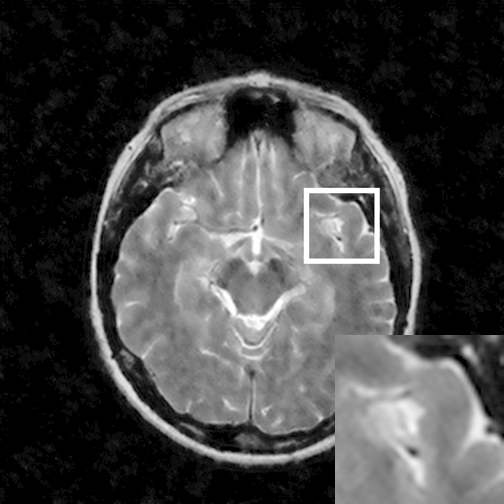}}
\ \ \ \ \ \subfigure[U-net]
{\includegraphics[height=1in,width=1in,angle=0]{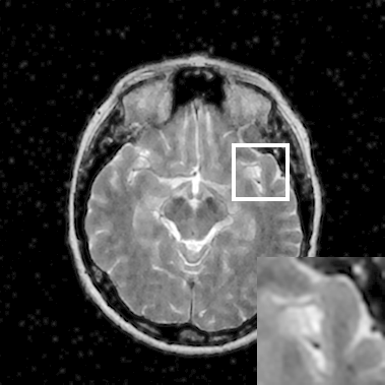}}
\ \ \ \ \ \subfigure[MDN]
{\includegraphics[height=1in,width=1in,angle=0]{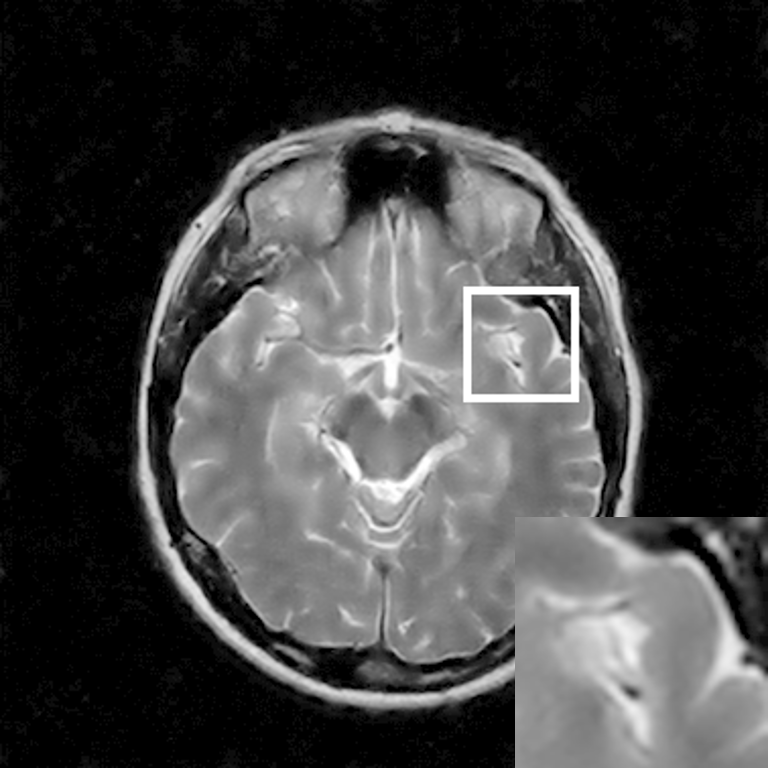}}\\
\subfigure[\tiny Sparse MRI]
{\includegraphics[height=0.7in,width=0.7in,angle=0]{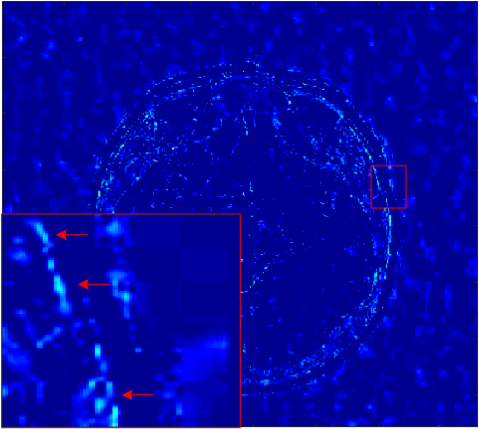}}
\ \ \subfigure[\tiny DLMRI]
{\includegraphics[height=0.7in,width=0.7in,angle=0]{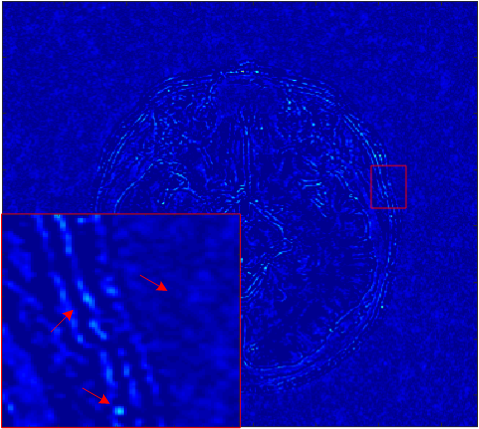}}
\ \ \subfigure[\tiny Single-scale]
{\includegraphics[height=0.7in,width=0.7in,angle=0]{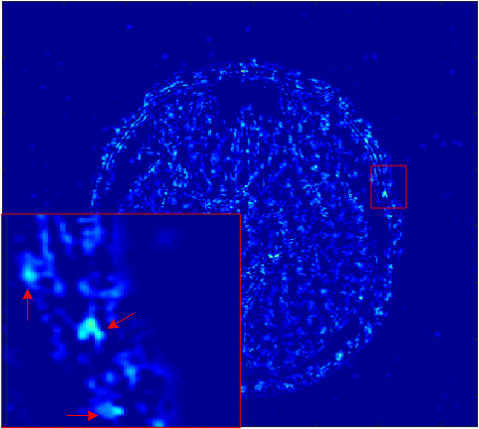}}
\ \ \subfigure[\tiny LRLs]
{\includegraphics[height=0.7in,width=0.7in,angle=0]{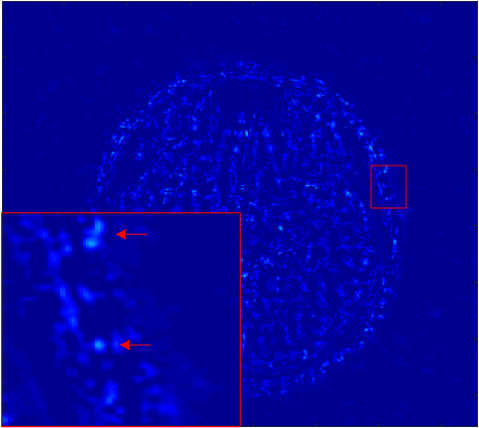}}
\ \ \subfigure[\tiny U-net]
{\includegraphics[height=0.7in,width=0.7in,angle=0]{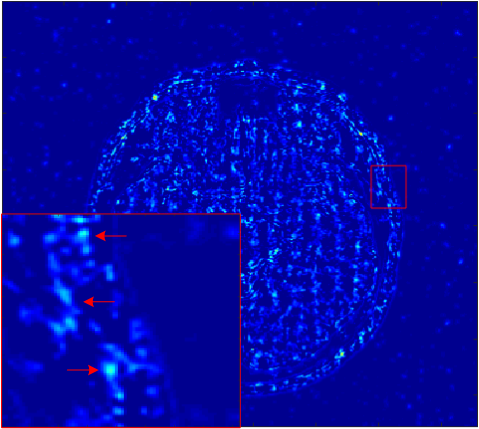}}
\ \subfigure[\tiny MDN]
{\includegraphics[height=0.7in,width=0.7in,angle=0]{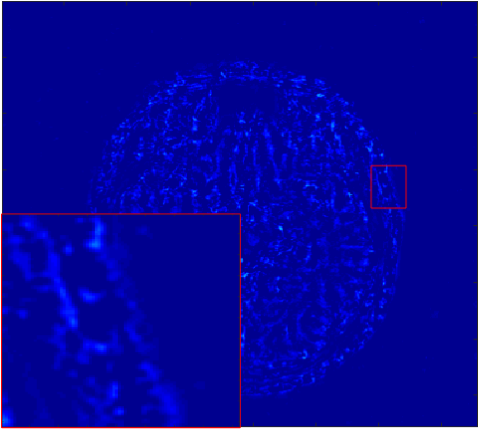}}
\caption{Reconstruction results for 20\% variable density sampling. (a) Original. (b)-(h) Reconstructed images. (i)-(n) The errors of six CSMRI methods. }
\end{figure*}
\begin{figure*}[!t]
\graphicspath{{figure15/}}
\centering
\subfigure[Ground truth]
{\includegraphics[height=1in,width=1in,angle=0]{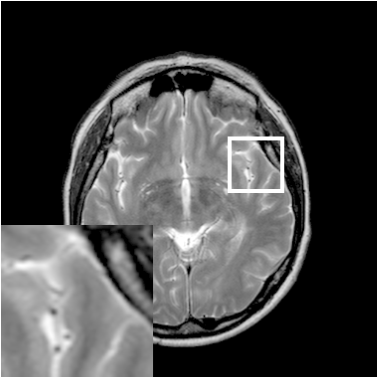}}
\ \ \ \ \ \subfigure[Zero-filling]
{\includegraphics[height=1in,width=1in,angle=0]{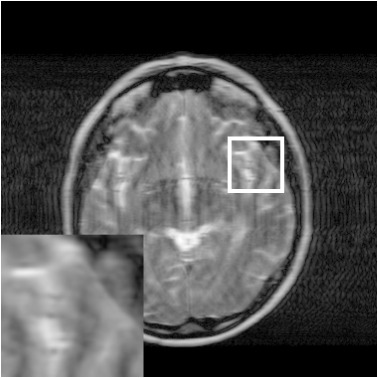}}
\ \ \ \ \ \subfigure[Sparse MRI]
{\includegraphics[height=1in,width=1in,angle=0]{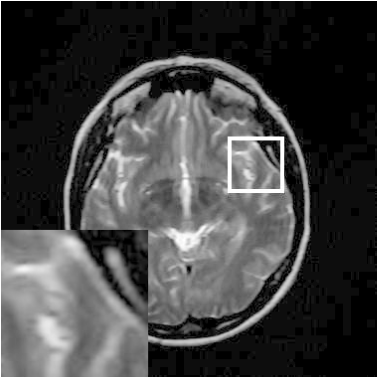}}
\ \ \ \ \ \subfigure[DLMRI]
{\includegraphics[height=1in,width=1in,angle=0]{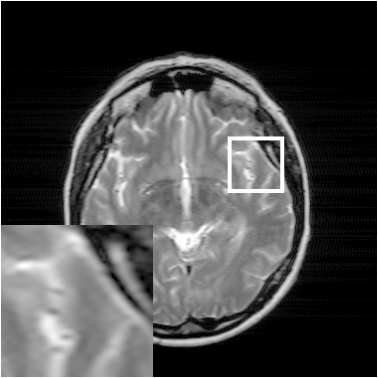}}
\ \ \ \ \ \subfigure[Single-scale]
{\includegraphics[height=1in,width=1in,angle=0]{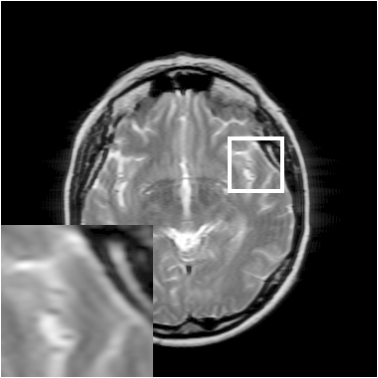}}
\ \ \ \ \ \subfigure[LRLs]
{\includegraphics[height=1in,width=1in,angle=0]{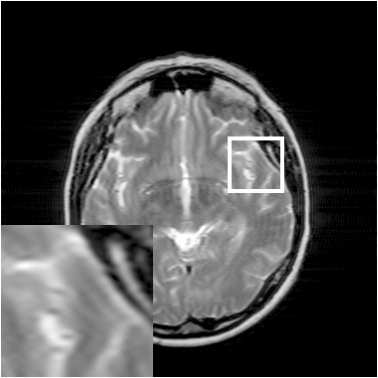}}
\ \ \ \ \ \subfigure[U-net]
{\includegraphics[height=1in,width=1in,angle=0]{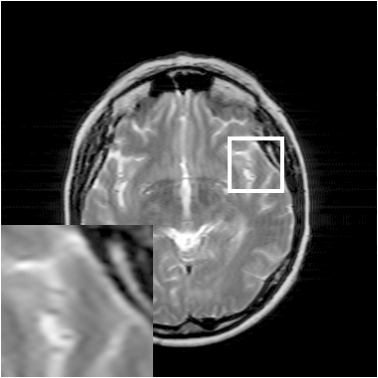}}
\ \ \ \ \ \subfigure[MDN]
{\includegraphics[height=1in,width=1in,angle=0]{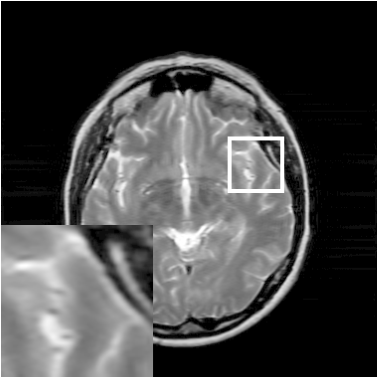}}\\
\subfigure[\tiny Sparse MRI]
{\includegraphics[height=0.7in,width=0.7in,angle=0]{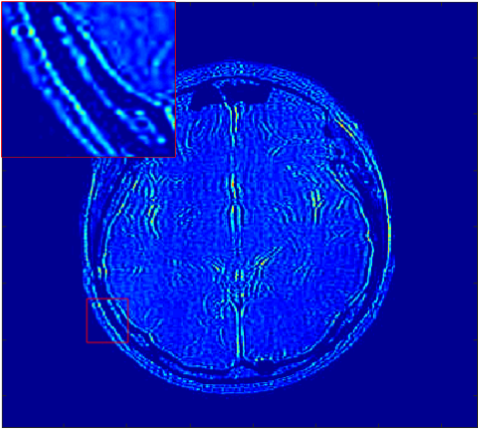}}
\ \ \subfigure[\tiny DLMRI]
{\includegraphics[height=0.7in,width=0.7in,angle=0]{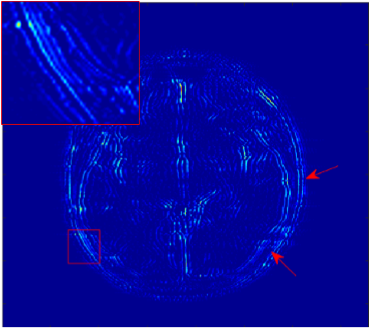}}
\ \ \subfigure[\tiny Single-scale]
{\includegraphics[height=0.7in,width=0.7in,angle=0]{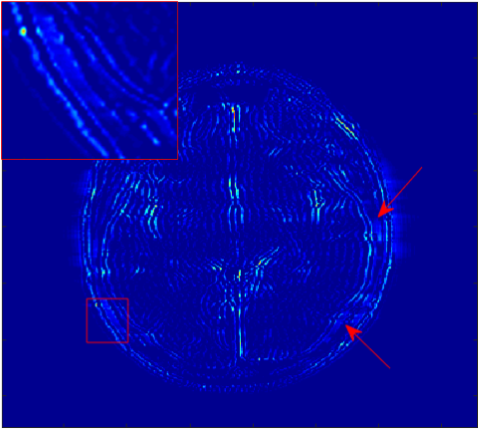}}
\ \ \subfigure[\tiny LRLs]
{\includegraphics[height=0.7in,width=0.7in,angle=0]{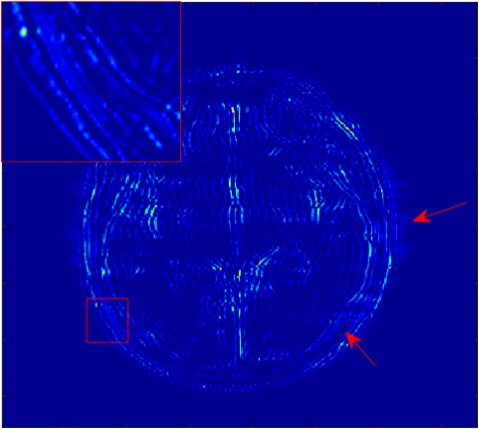}}
\ \ \subfigure[\tiny U-net]
{\includegraphics[height=0.7in,width=0.7in,angle=0]{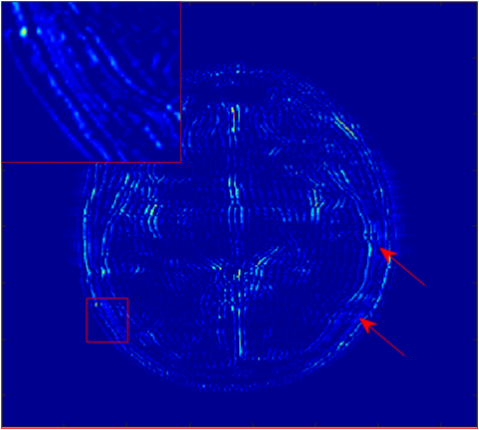}}
\ \subfigure[\tiny MDN]
{\includegraphics[height=0.7in,width=0.7in,angle=0]{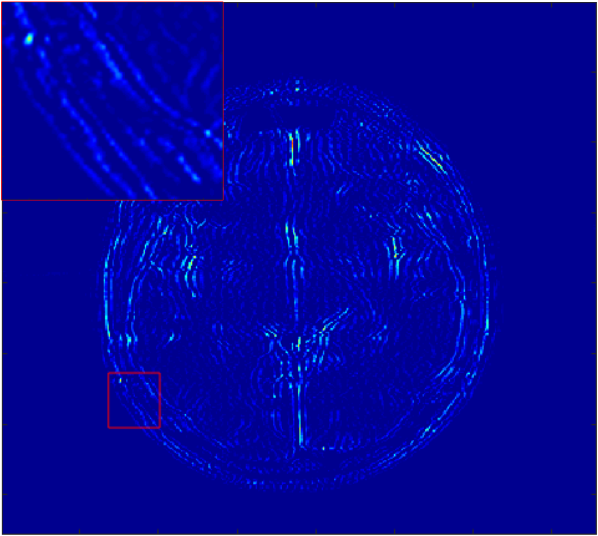}}
\caption{Reconstruction results for 25\% cartesian sampling. (a) Original. (b)-(h) Reconstructed images. (i)-(n) The errors of six CSMRI methods.}
\end{figure*}
\begin{figure*}[!t]
\graphicspath{{figure15/}}
\centering
\subfigure[Ground truth]
{\includegraphics[height=1in,width=1in,angle=0]{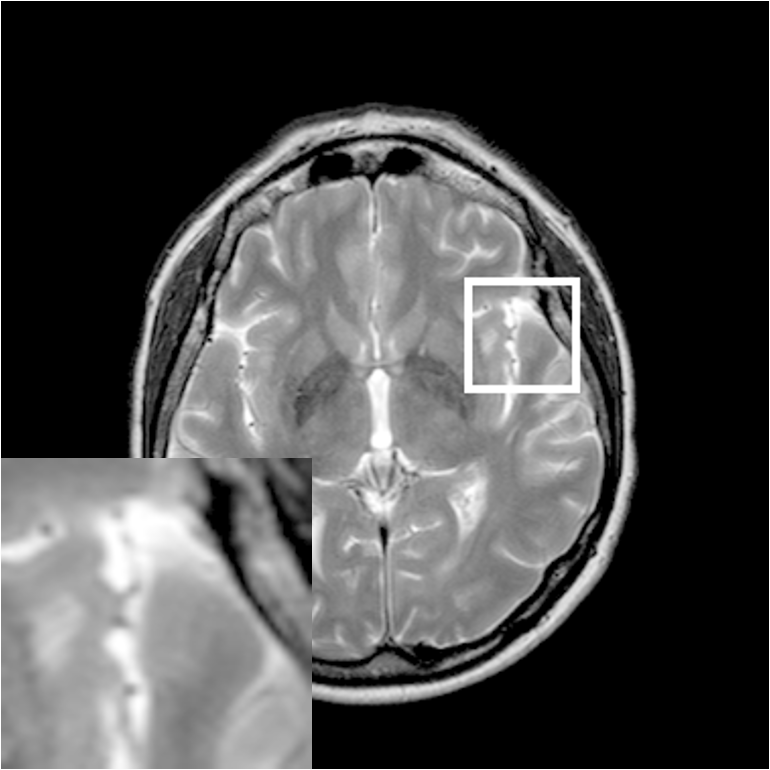}}
\ \ \ \ \ \subfigure[Zero-filling]
{\includegraphics[height=1in,width=1in,angle=0]{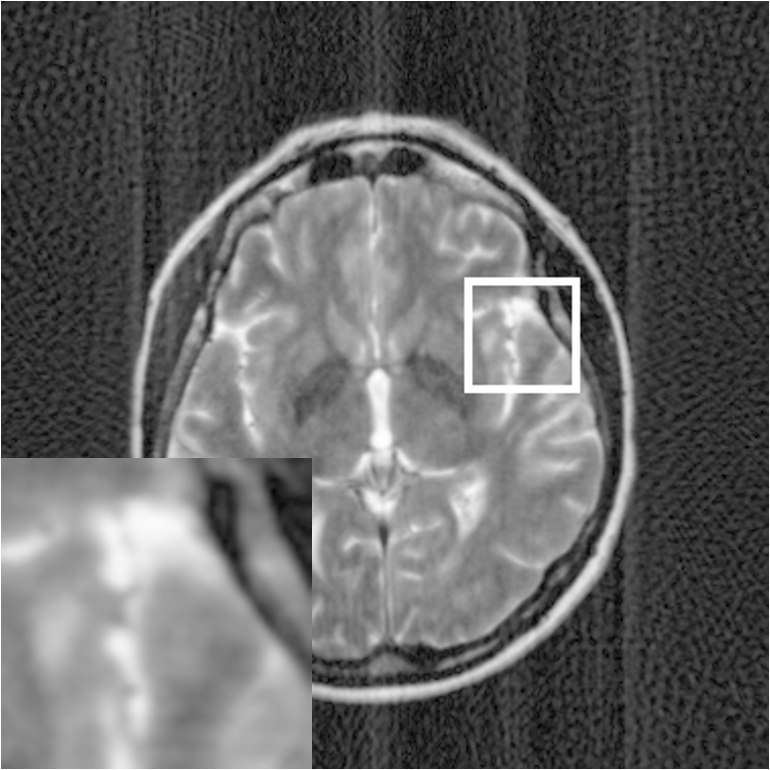}}
\ \ \ \ \ \subfigure[Sparse MRI]
{\includegraphics[height=1in,width=1in,angle=0]{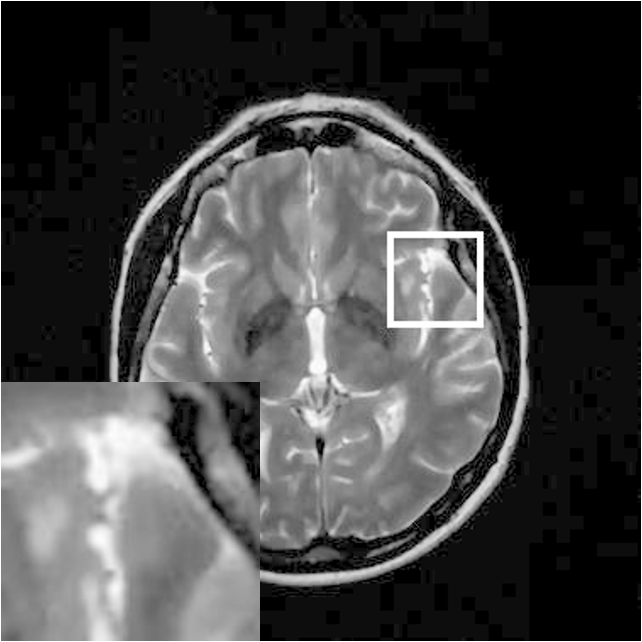}}
\ \ \ \ \ \subfigure[DLMRI]
{\includegraphics[height=1in,width=1in,angle=0]{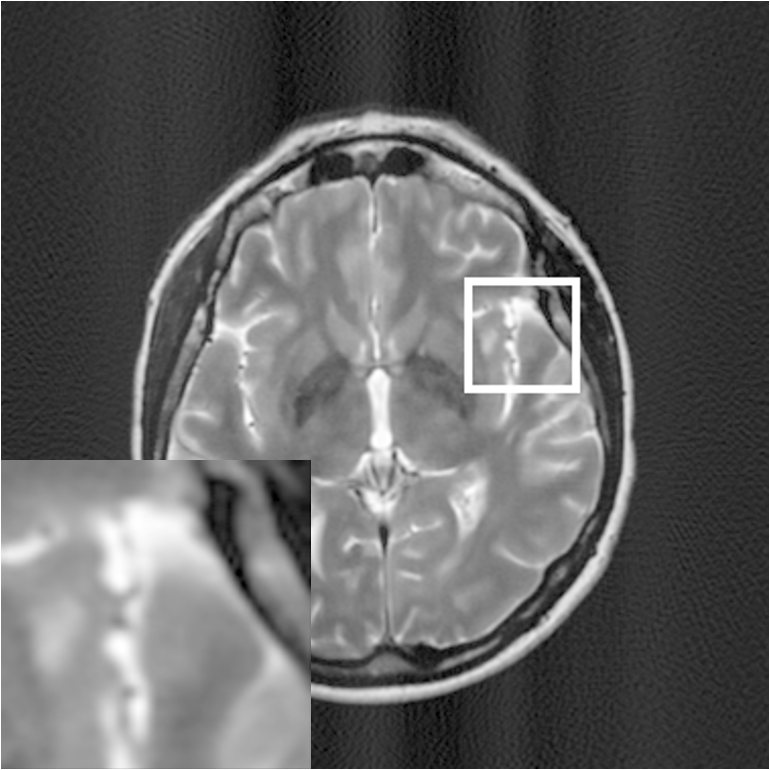}}
\ \ \ \ \ \subfigure[Single-scale]
{\includegraphics[height=1in,width=1in,angle=0]{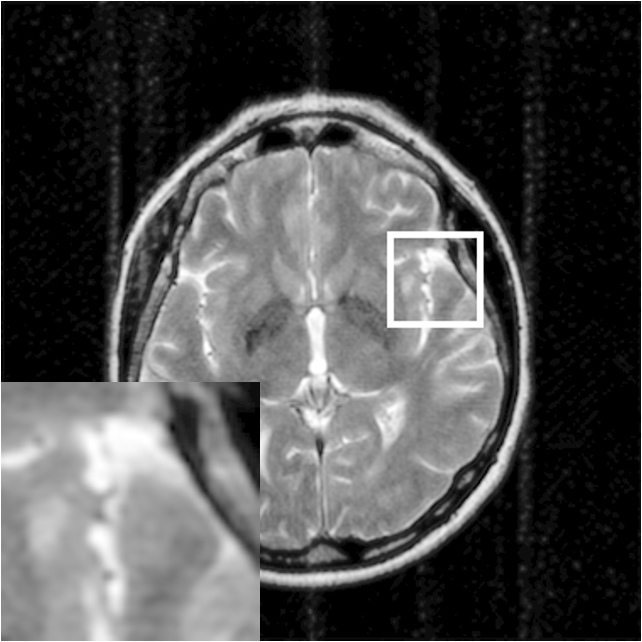}}
\ \ \ \ \ \subfigure[LRLs]
{\includegraphics[height=1in,width=1in,angle=0]{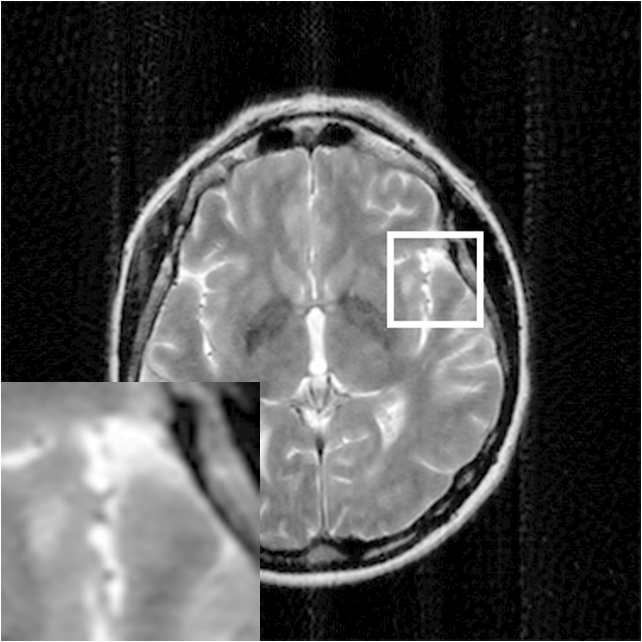}}
\ \ \ \ \ \subfigure[U-net]
{\includegraphics[height=1in,width=1in,angle=0]{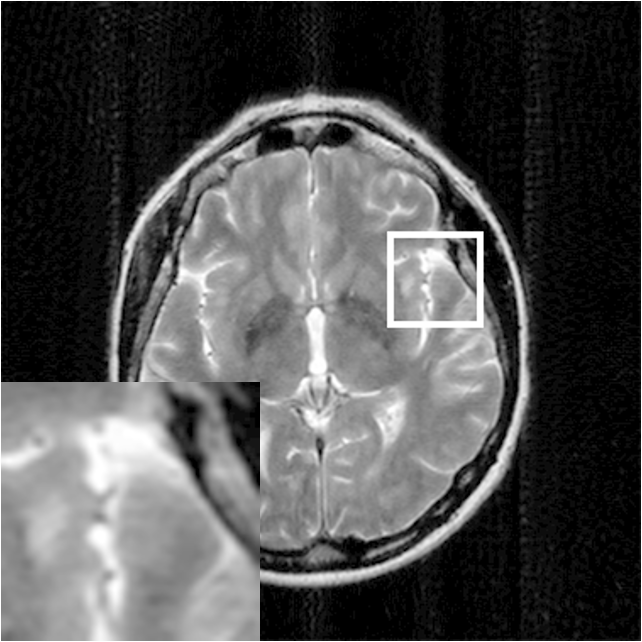}}
\ \ \ \ \ \subfigure[MDN]
{\includegraphics[height=1in,width=1in,angle=0]{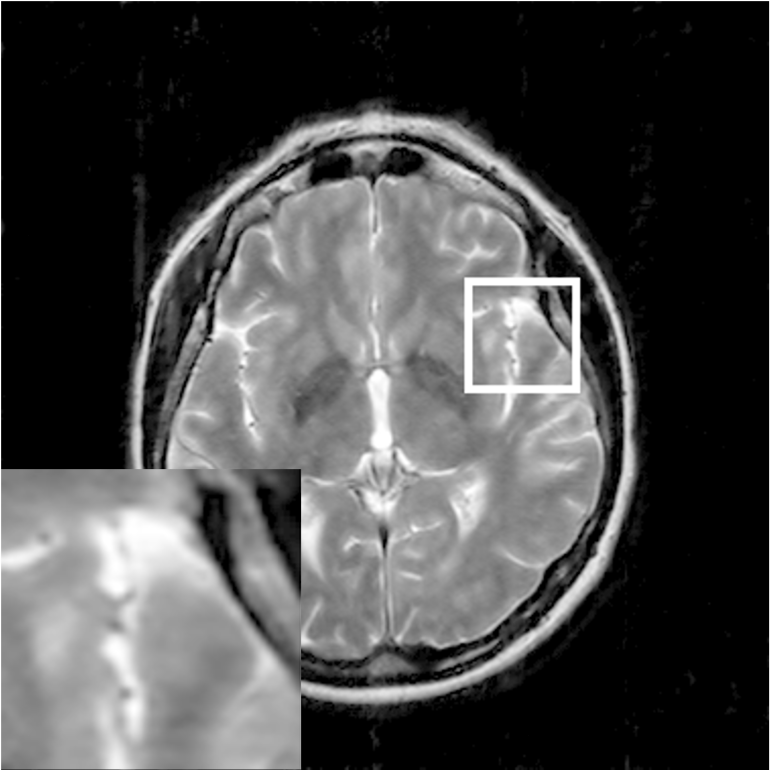}}\\
\subfigure[\tiny Sparse MRI]
{\includegraphics[height=0.7in,width=0.7in,angle=0]{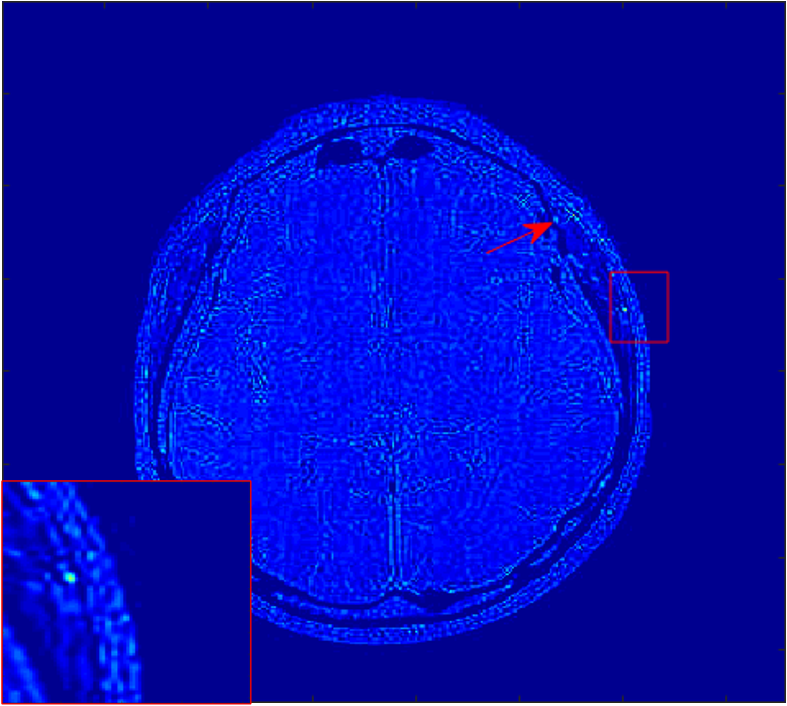}}
\ \ \subfigure[\tiny DLMRI]
{\includegraphics[height=0.7in,width=0.7in,angle=0]{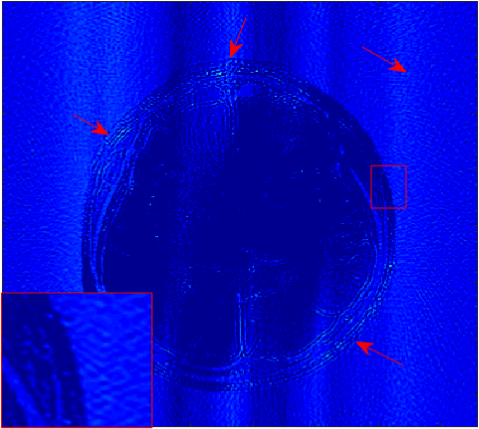}}
\ \ \subfigure[\tiny Single-scale]
{\includegraphics[height=0.7in,width=0.7in,angle=0]{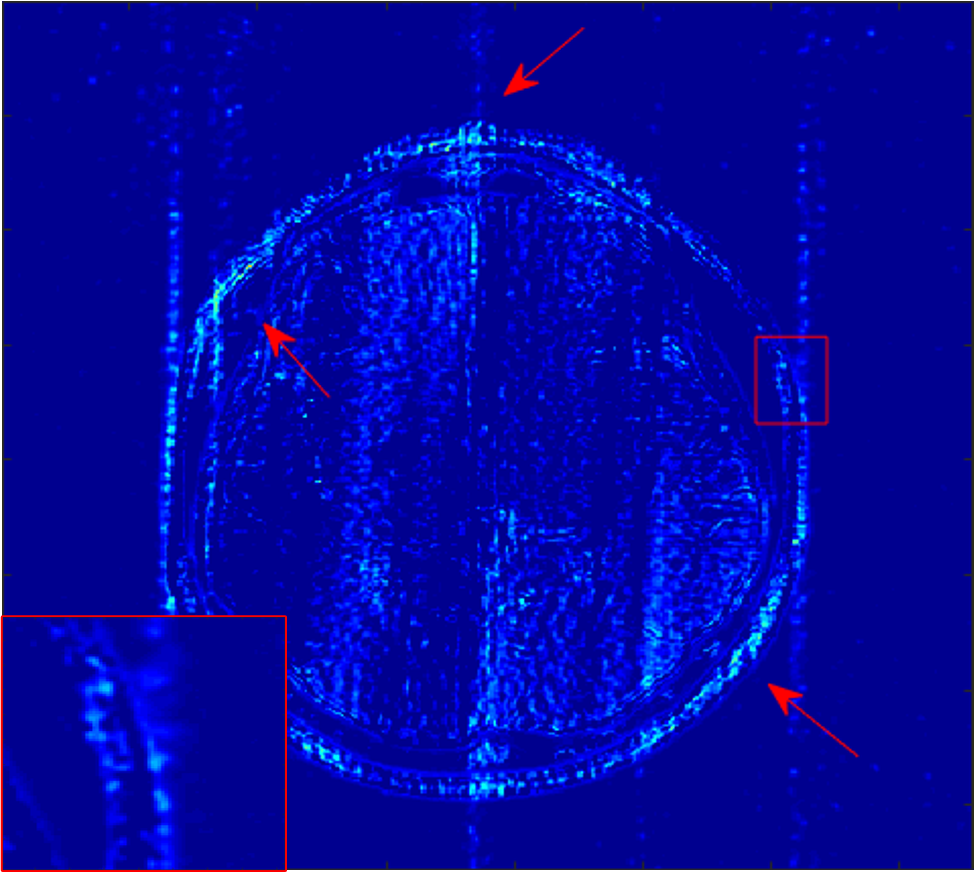}}
\ \ \subfigure[\tiny LRLs]
{\includegraphics[height=0.7in,width=0.7in,angle=0]{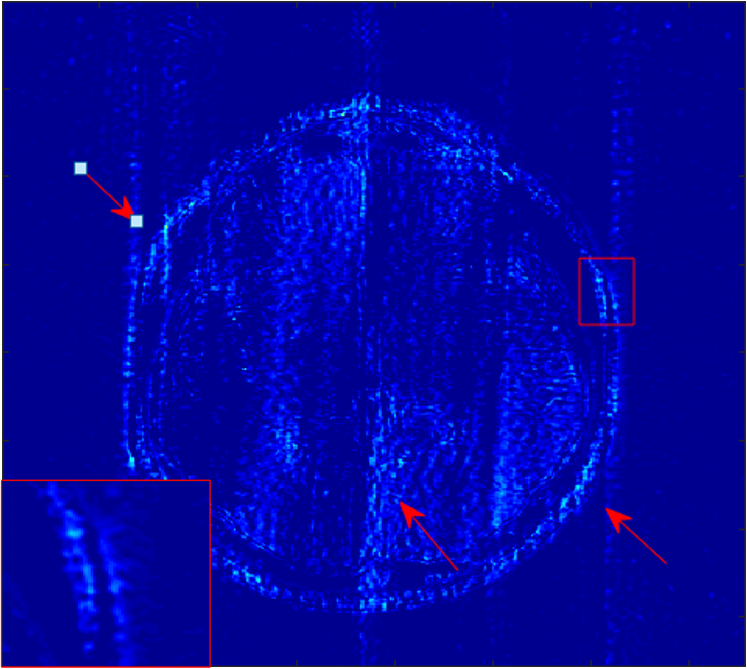}}
\ \ \subfigure[\tiny U-net]
{\includegraphics[height=0.7in,width=0.7in,angle=0]{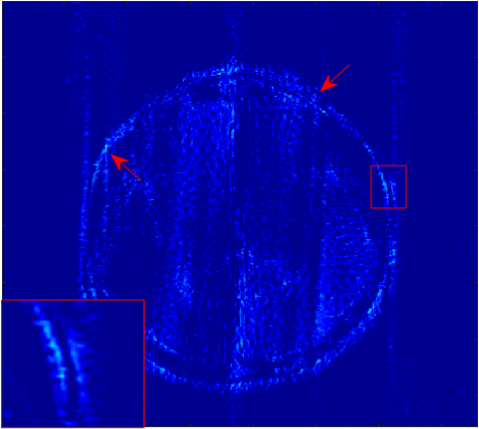}}
\ \subfigure[\tiny MDN]
{\includegraphics[height=0.7in,width=0.7in,angle=0]{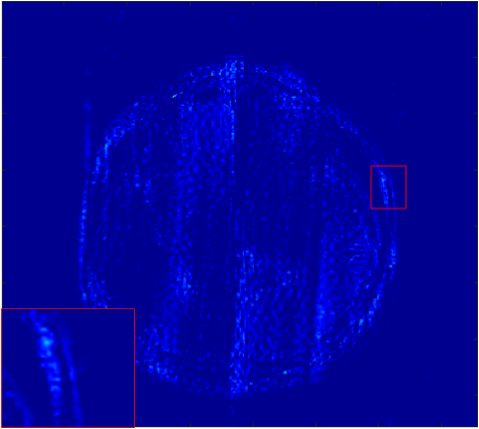}}
\caption{Reconstruction results for 30\% radial sampling. (a) Original. (b)-(h) Reconstructed images. (i)-(n) The errors of six CSMRI methods.}
\end{figure*}

\begin{figure*}[!t]
\graphicspath{{figure15/}}
\centering
\subfigure[Cartesian sampling]
{\includegraphics[height=1.7in,width=2.2in,angle=0]{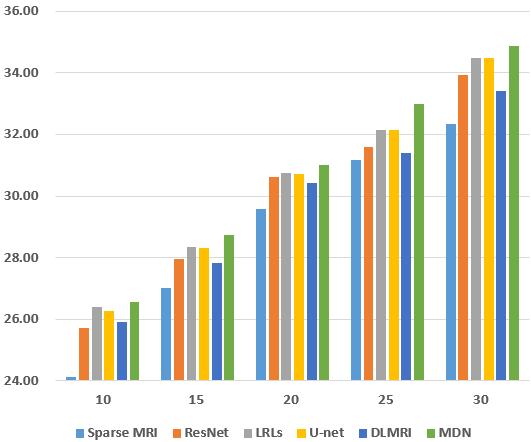}}
\ \ \ \ \subfigure[Random sampling]
{\includegraphics[height=1.7in,width=2.2in,angle=0]{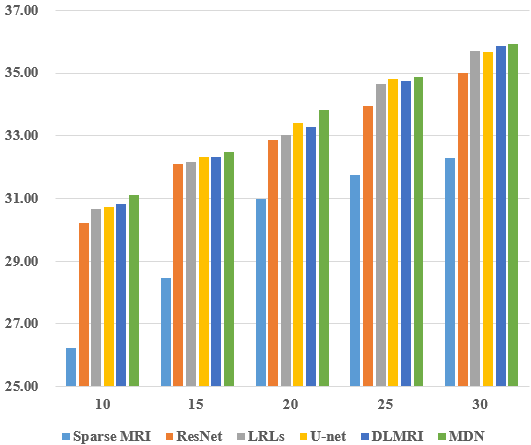}}
\ \ \ \ \subfigure[Radial sampling]
{\includegraphics[height=1.7in,width=2.2in,angle=0]{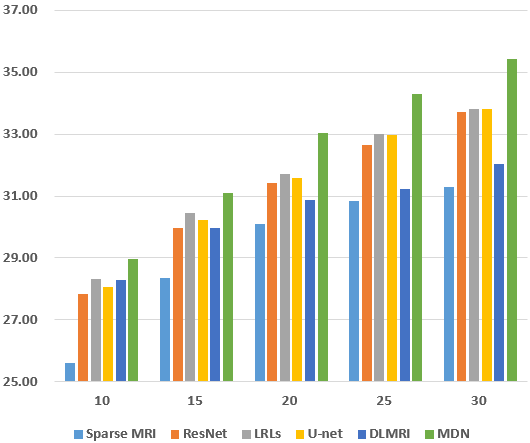}}
\ \ \ \ \subfigure[Standard deviation]
{\includegraphics[height=1.7in,width=2.2in,angle=0]{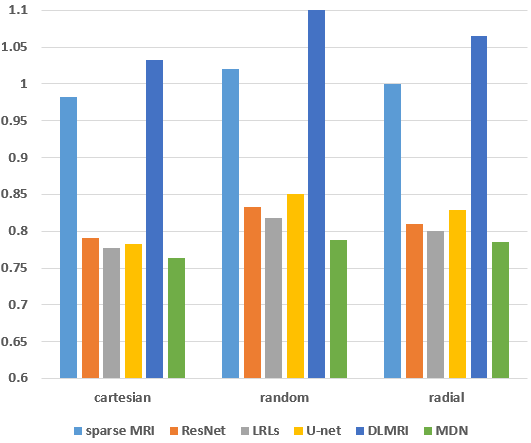}}
\subfigure[\tiny Sparse MRI]
{\includegraphics[height=0.7in,width=0.7in,angle=0]{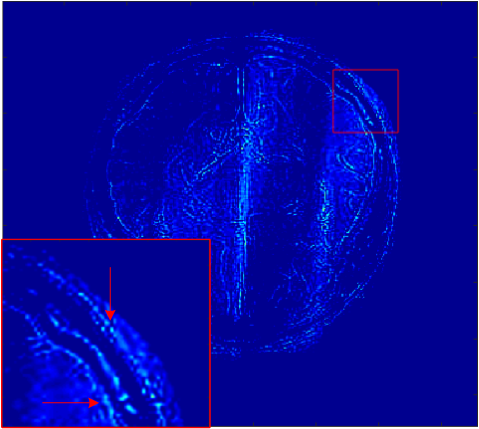}}
\ \ \subfigure[\tiny DLMRI]
{\includegraphics[height=0.7in,width=0.7in,angle=0]{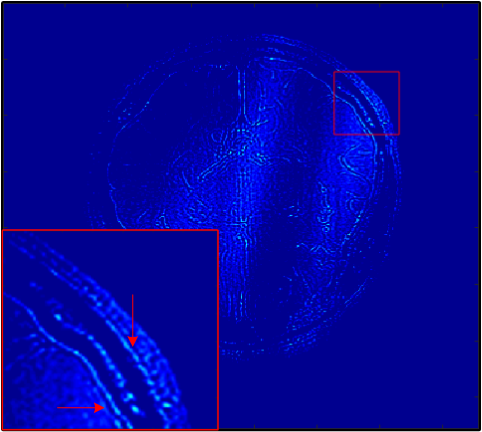}}
\ \ \subfigure[\tiny Single-scale]
{\includegraphics[height=0.7in,width=0.7in,angle=0]{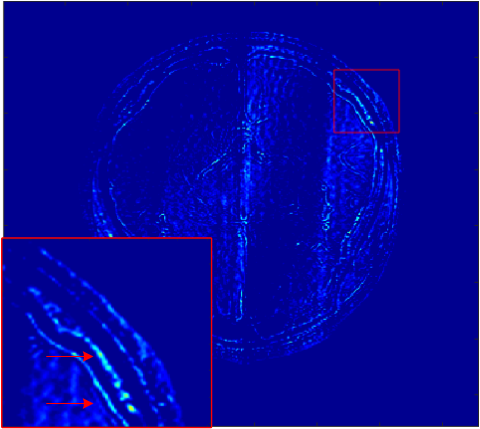}}
\ \ \subfigure[\tiny LRLs]
{\includegraphics[height=0.7in,width=0.7in,angle=0]{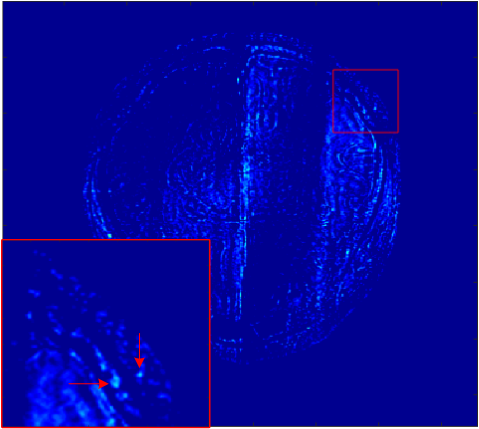}}
\ \ \subfigure[\tiny U-net]
{\includegraphics[height=0.7in,width=0.7in,angle=0]{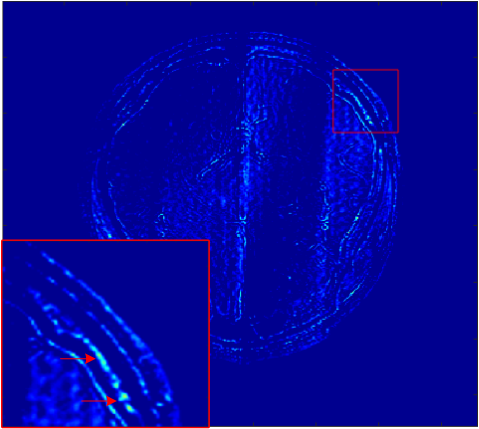}}
\ \subfigure[\tiny MDN]
{\includegraphics[height=0.7in,width=0.7in,angle=0]{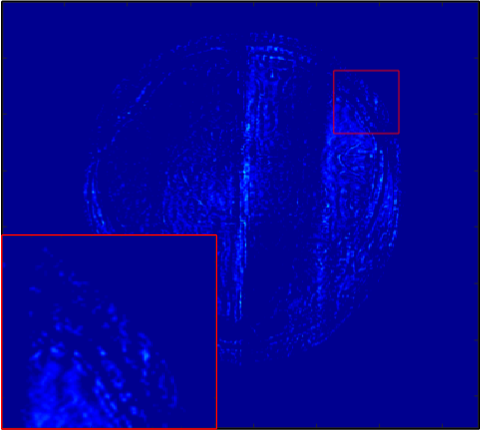}}
\caption{The reconstruction results on complex-valued MRI data. (a)-(c) are the PSNR values of different methods with three sampling masks and five sampling rates, in which the x-axis represents sampling rates. (d) is the standard deviation of PSNR values on different methods when using 30\% sampling rates, in which the x-axis represents sampling masks. And (e)-(j) are the errors of six CSMRI methods.}
\end{figure*}

As shown in Figs. 7, 8 and 9, Sparse MRI and DLMRI have a lot of unpleasant artifacts, Residual learning and U-net can eliminate most of artifacts, but are not ideal for restoring image details. However, the proposed method can reconstruct better MR images, which outperforms other competitive methods in visualization of structures reconstruction and artifacts removal. Meanwhile, we can see from the absolute error residuals for three sampling experiments that the proposed MDN algorithm restores a finer detail structure than other algorithms. Moreover, we present the PSNR and SSIM values in Table I for different algorithms, sampling masks and sampling rates. It is demonstrated that the proposed method provides better reconstruction performance and visual results than other competitive methods. We can also see the obvious improvement of all algorithms over zero-filling both in visualization. In particular, a higher SSIM value of Sparse MRI appears when using 30\% variable density random sampling, however, Sparse MRI generates more artifacts than the proposed MDN.

\begin{table}[]
\centering
\caption{\label{tab:results} PSNR/SSIM/training time for the different combinations of $N_f$ and \emph{DF} from 2-nd layer to 7-th layer. The first layer is fixed: $N_f$=32, \emph{DF}=1. The value with red bold font indicates ranking the first place in this column while values with blue font are the second and third place.}
\begin{tabular}{|c|l|c|r|c|}
\hline
$N_f$                                      & \multicolumn{1}{c|}{DF} & PSNR  & SSIM  & Training time (mins) \\ \hline
\multirow{4}{*}{64$\times$6}                     & 3-3-3-3-3-3             & 34.64 & \textcolor[rgb]{1.00,0.00,0.00}{0.946} & 782.5         \\ \cline{2-5}
                                        & 2-2-2-2-2-2             & \textcolor[rgb]{1.00,0.00,0.00}{34.98} & 0.940 & 720           \\ \cline{2-5}
                                        & 2-3-2-3-2-3             & 34.88 & \textcolor[rgb]{0.00,0.07,1.00}{0.945} & 752.5         \\ \cline{2-5}
                                        & 3-2-3-2-3-2             & \textcolor[rgb]{0.00,0.07,1.00}{34.97} & 0.937 & 752.5         \\ \hline
\multirow{4}{*}{32$\times$6}                     & 3-3-3-3-3-3             & 34.85 & 0.940 & 685           \\ \cline{2-5}
                                        & 2-2-2-2-2-2             & 34.62 & 0.931 & 645           \\ \cline{2-5}
                                        & 2-3-2-3-2-3             & 34.83 & 0.930 & 662.5         \\ \cline{2-5}
                                        & 3-2-3-2-3-2             & 34.83 & 0.931 & 667.5         \\ \hline
\multicolumn{1}{|l|}{\textbf{64-32-64-32-64-32}} & \textbf{2-3-2-3-2-3}             & \textcolor[rgb]{0.00,0.07,1.00}{\textbf{34.95}} & \textcolor[rgb]{0.00,0.07,1.00}{\textbf{0.944}} & \textbf{700}           \\ \hline
\end{tabular}
\end{table}
\begin{table*}[t]
\centering
 \begin{threeparttable}
 {\caption{\label{tab:results} PSNR/SSIM for the ablation study on concatenation layers with different sampling masks and specific sampling rates.}
  \setlength{\tabcolsep}{3mm}{
  \begin{tabular}{cccccccc}
  \toprule
  Mask $\&$ sampling rate & Random $\&$ 20\% & Cartesian $\&$ 25\% & Radial $\&$ 30\% \\
  \midrule
  no concat &33.54/0.903 &32.49/0.923 &34.41/0.872 \\
  \textbf{with concat} &\textbf{34.95/0.944} &\textbf{33.25/0.950} &\textbf{35.64/0.955} \\
  \bottomrule
  \end{tabular}}}
 \end{threeparttable}
\end{table*}

\begin{table*}[t]
\centering
 \begin{threeparttable}
 \caption{\label{tab:results} PSNR/SSIM for the ablation study about residual learnings based on dilated convolutions.}
  \setlength{\tabcolsep}{3mm}{
  \begin{tabular}{cccccccc}
  \toprule
  Learning rate & 0.0001 & 0.001 & 0.01 \\
  \midrule
  \textbf{GRL $\surd$  LRLs $\surd$ } &\textbf{34.95/0.935} &\textbf{34.95/0.944} &\textbf{34.31/0.939} \\
  GRL $\times$ LRLs $\surd$ &34.15/0.912 &34.17/0.919 &33.74/0.924 \\
  GRL $\surd$ LRLs $\times$ &34.31/0.905 &34.52/0.902 &31.61/0.662 \\
  GRL $\times$ LRLs $\times$ &32.00/0.781 &31.61/0.702 &33.56/0.584 \\
  \bottomrule
  \end{tabular}}
 \end{threeparttable}
\end{table*}
\subsection{\textbf{Experiments on complex-valued MRI with different masks}}
We evaluate the performance of the proposed model using PSNR on complex-valued data and compare with two optimization-based methods and three deep-learning methods. We present the PSNR results for all sampling masks and five rates in Figs. 10(a)-(c) and it is obvious that the proposed model outperforms other five methods, which can demonstrate the effectiveness of MDN model on complex-valued data. Additionally, we provide the standard deviation on 80 test images of different methods when using 30\% sampling rates of three masks in Fig. 10(d). We can observe that deep-learning methods obtain more stable performance than DLMRI and Sparse MRI. In Figs. 10(e)-(j), we show the absolute value of residuals of different algorithms using 30\% radial sampling rate. We can see that the proposed model has less noise-like errors than other five methods.
\subsection{\textbf{Ablation Study}}
 {\textbf{Ablation study on network size setting.} As mentioned above, we choose proper \emph{DF} and ${N_f}$ under the consideration of network size and performance. We conduct several experiments about the setting of \emph{DF} and ${N_f}$ in Table 2 and demonstrate the PSNR/SSIM values of different combinations. Additionally, we show the training time to evaluate computational cost with various network sizes.}

 In MDN blocks, the first layer with 9$\times$9 kernel aims to enlarge receptive fields to extract more initial information for the block with no necessary to employ larger \emph{DF} and $N_f$. We make a comparison between 9$\times$9 kernel with 32 filters and 3$\times$3 kernel with 64 filters in the first layer, and the former increases the value of PSNR by 0.1 than the latter. Therefore, we fix the first layer as shown in Fig. 4. In Table 2, all channels (${N_f}$) of feature maps in MDN blocks set to 64 indeed increases the training time with a little improvement in PSNR/SSIM, however, all ${N_f}$ set to 32 decreases reconstruction results in despite of less training time. Considering training time, reconstruction results and application of local residual learnings, we choose the alternating ${N_f}$ values of 64 and 32. Meanwhile, we employ larger \emph{DF} values for the layers with 32 feature maps in order to supplement some useful information extracted by enlarged receptive fields. By the way, setting larger \emph{DF} than 3 obviously burdens the network and increases the training time.

 {\textbf{Ablation study on the concat layer.} To demonstrate the effectiveness of fusing multi-scale features, we conduct the ablation investigation on concatenation layers. It can be noticeable in Table 3 that using concat layers to fuse multi-scale features extracted from dilated network can achieve better reconstruction.}

 {\textbf{Ablation study on residual learnings and investigation on initial learning rates.} We have explained that the proposed MDN integrates GRL and LRLs to maintain the abundance of feature maps for better reconstruction. And in this section, we show the results in terms of PSNR and SSIM among non-residual, global residual, local residual and MDN, in which all of them are based on multi-scale dilated network. As shown in Table 4, MDN which integrates GRL and LRLs outperforms other residual learnings and non-residual learning. It is obvious that MDN extracts more valid feature maps which can provide better reconstruction. Based on residual experiments, we consider the effort of different initial learning rates on reconstruction as well. It can be noticed in Table 4 that the four networks generally perform outstandingly in 0.001. As a consequence, we set initial learning rate as 0.001 during all training process.}

\subsection{\textbf{Experiments in the noisy setting}}

\begin{table*}[t]
\centering
 \begin{threeparttable}
 \caption{\label{tab:results} PSNR for 35\% variable density sampling of brain MR with various noise standard deviations}
  \setlength{\tabcolsep}{3mm}{
  \begin{tabular}{cccccccc}
  \toprule
  Reconstruction method & v=0 & v=0.01 & v=0.02 &v=0.03 \\
  \midrule
  Zero-filled $\&$ noisy &31.50 &20.88 &18.34 &16.81 \\
  LRLs &37.53 &31.09 &29.78 &29.22 \\
  \textbf{MDN} &\textbf{38.06} &\textbf{31.74} &\textbf{30.36} &\textbf{29.54} \\

  \bottomrule
  \end{tabular}}
 \end{threeparttable}
\end{table*}
\begin{figure*}[!t]
\graphicspath{{figure15/}}
\centering
\ \ \ \ \ \ \ \subfigure[Zero-filled $\&$ noisy]
{\includegraphics[height=1.2in,width=1.2in,angle=0]{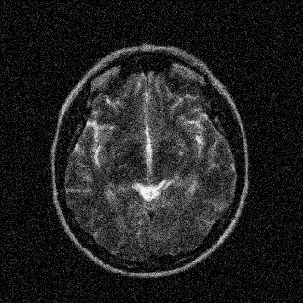}}
\ \ \ \ \ \ \ \subfigure[LRLs]
{\includegraphics[height=1.2in,width=1.2in,angle=0]{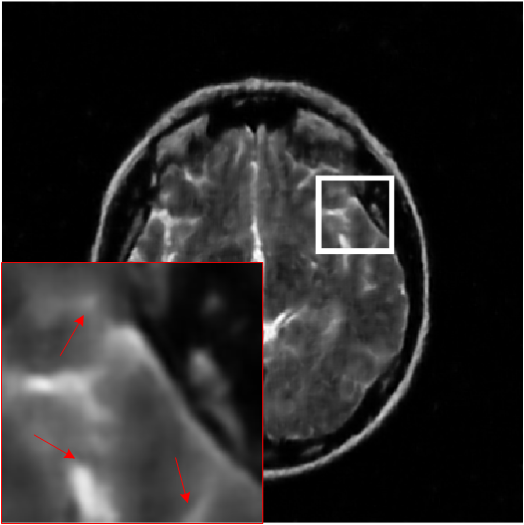}}
\ \ \ \ \ \ \ \subfigure[MDN]
{\includegraphics[height=1.2in,width=1.2in,angle=0]{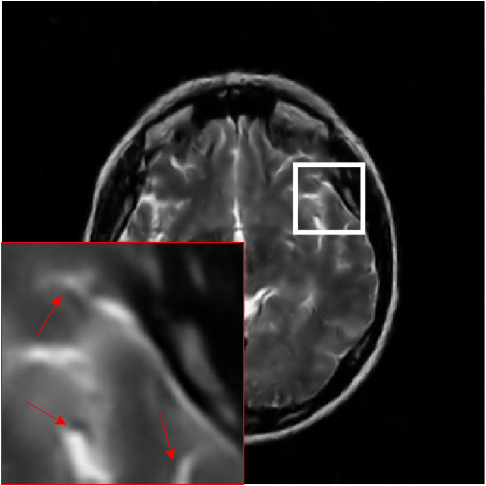}}
\caption{Reconstruction for noisy images based on zero-filling, in which the noise standard deviation is set to 0.01.}
\end{figure*}

The MR imaging we considered above have been completely noiseless. However, unexpected noise may be mixed in the sampling process for some external conditions. We continue our evaluation of noisy MRI to verify the stability of reconstruction based on MDN. Moreover, we compare the proposed MDN with one deep learning-based algorithm (LRLs) in terms of visualization and metrics. The noisy MR images are respectively mixed with complex white Gaussian noise having standard deviation v = 0.01, 0.02, 0.03. And ground truth is the original noise-free MRI. It can be noted that the proposed MDN achieves better results than the LRLs method in terms of PSNR in Table 5. Fig. 11 shows the reconstruction based on MDN, in which the noisy image has been well recovered.

\subsection{\textbf{Discussions on dilated convolutions, the number of blocks and parameters.}}
\begin{table*}[t]
\centering
 \begin{threeparttable}
 \caption{\label{tab:results} PSNR/SSIM for non-dilated and dilated networks with several blocks}
  \setlength{\tabcolsep}{3mm}{
  \begin{tabular}{cccccccc}
  \toprule
  Block & 1 & 2 & 3 \\
  \midrule
  non-dilated &34.68/0.917 &34.78/0.940 &33.95/0.919 \\
  \textbf{dilated} &\textbf{34.85/0.930} &\textbf{34.95/0.944} &\textbf{34.98/0.934} \\
  \bottomrule
  \end{tabular}}
 \end{threeparttable}
\end{table*}
\begin{figure*}[!t]
\graphicspath{{figure15/}}
\centering
\ \ \ \subfigure[Number of blocks vs number of parameters.]
{\includegraphics[height=1.6in,width=2.1in,angle=0]{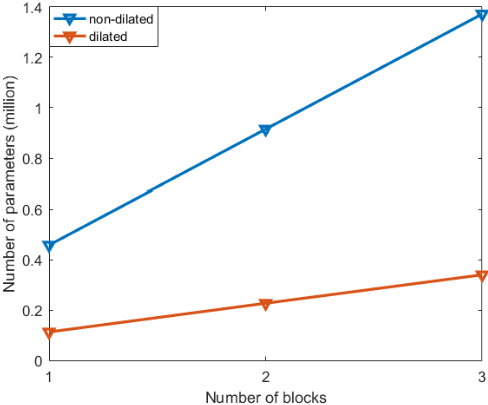}}
\ \ \ \subfigure[Deep learning methods vs number of parameters]
{\includegraphics[height=1.6in,width=2.1in,angle=0]{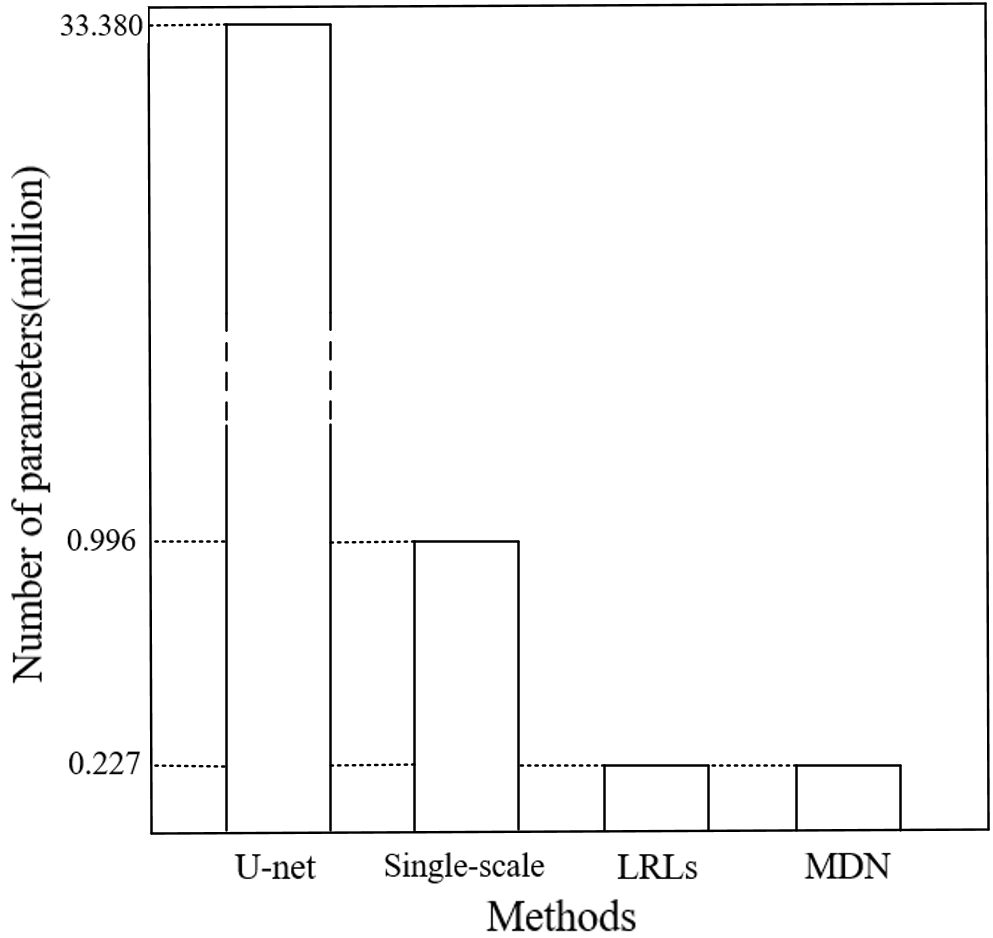}}
\caption{{The number of parameters.}}
\end{figure*}
We also verify the effect of the number of MDN-blocks, and calculate the number of correspond parameters, which aims to obtain a tradeoff between network size and performance. For non-dilated network, we control the receptive fields of convolutions consistent with dilated convolution. For the receptive field, 2-dilated $\&$ 3$\times$3 convolution is equivalent to non-dilated $\&$ 5$\times$5 convolution; and 3-dilated $\&$ 3$\times$3 convolution is equivalent to non-dilated $\&$ 7$\times$7 convolution. From the results of Table 6 and Fig. 12 referring to parameters calculation, two dilated blocks perform better reconstruction with less parameters. And it is obvious that the MDN achieves better reconstruction results with least parameters than other deep learning methods. As a consequence, the number of blocks should be set to 2, which can perform better results with a guarantee of training speed for the huge data sets.

\subsection{\textbf{Experiments on super-resolution}}
\begin{figure*}[!t]
\graphicspath{{figure15/}}
\centering
\ \ \ \ \ \ \ \subfigure[MDN]
{\includegraphics[height=1.5in,width=1.5in,angle=0]{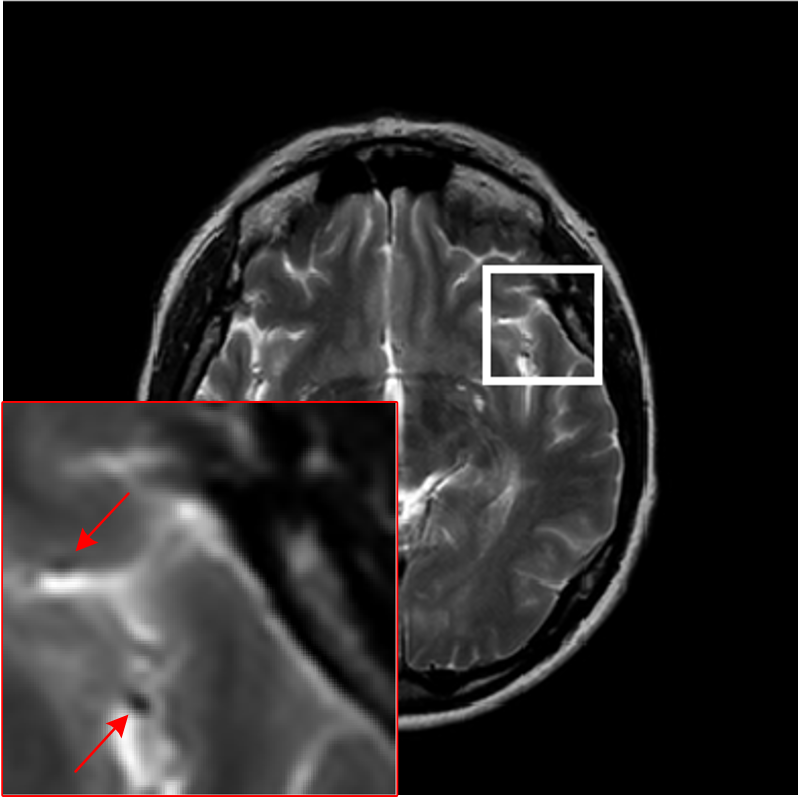}}
\ \ \ \ \ \ \ \subfigure[VDSR]
{\includegraphics[height=1.5in,width=1.5in,angle=0]{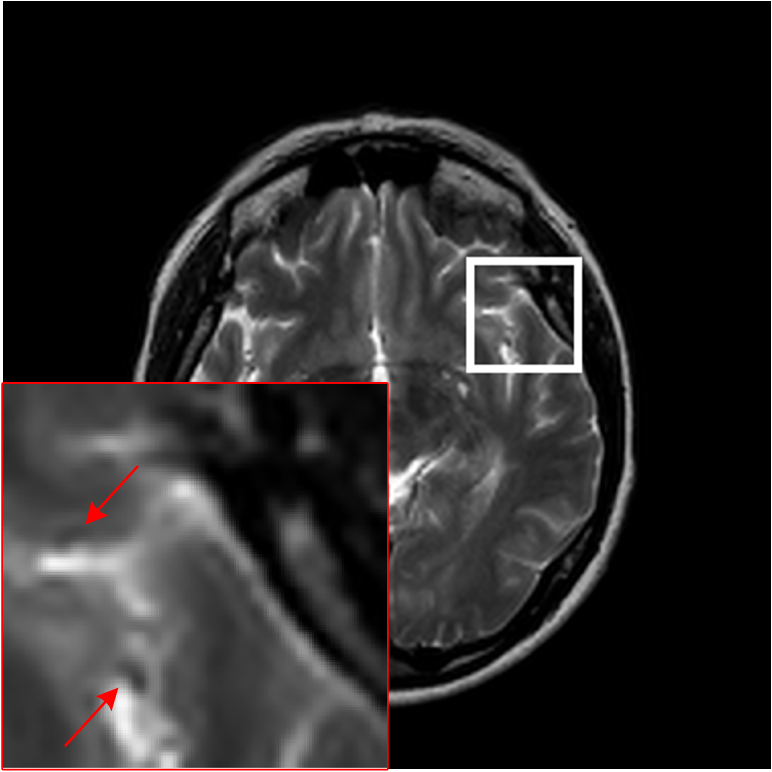}}
\caption{{Super-resolution results of a brain MRI with scale factor $\times$2.}}
\end{figure*}
\begin{table*}[t]
\centering
 \begin{threeparttable}
 \caption{\label{tab:results} PSNR/SSIM for scale factors $\times$2,$\times$3,$\times$4 on our data sets}
  \setlength{\tabcolsep}{1mm}{
  \begin{tabular}{cccccccc}
  \toprule
  Scale  & 2  & 3  & 4 \\
  \midrule
  \textbf{MDN} &\textbf{38.73/0.986} &\textbf{34.06/0.962} &\textbf{30.09/0.917} \\
  VDSR &38.04/0.983 &30.85/0.930 &29.60/0.906 \\
  difference &0.69/0.003 &3.21/0.032 &0.49/0.011\\
  \bottomrule
  \end{tabular}}
 \end{threeparttable}
\end{table*}
Subsequently, we conduct extended experiments on MR image super-resolution, which aims to recover high-resolution MR images from their low-resolution images for improving image analysis and visualization in the clinic. VDSR [24] trains a deep network with multiple scale factors for image super-resolution task which can reduce the number of parameters and achieve efficient results. We demonstrate the comparison results of the proposed MDN and VDSR in Table 7 and Fig. 13. It is noted that the proposed MDN performs better reconstruction results than VDSR on a huge dataset.

\section{CONCLUSION AND PROSPECT}
A novel multi-scale dilated network (MDN) has been presented for CSMRI. The proposed MDN is composed by cascading two basic blocks where dilated convolutions, global and local residual learnings, and concatenation layers are integrated to extend the receptive fields of convolutional kernels for reducing network parameters, maintaining features abundance, and fusing multi-scale features, respectively. Final experiments demonstrate that MDN achieves outstanding performance with training huge and diverse data, and the proposed network outperforms several competitive CSMRI algorithms in subjective and objective assessments. In addition, the proposed model is effective in MR noisy setting and super-resolution tasks.

In the future, we will adjust our model to parallel and dynamic imaging referred from [43] and [13]. And we will also improve our method with some variational models( [44] and [45] ) which is beneficial to image reconstruction. In addition to MR reconstruction, we will consider the application of our model in segmentation task [21].

\section{ACKNOWLEDGEMENT}
The authors sincerely thank anonymous editor and reviewers for their constructive and valuable comments. This work was supported in part by the National Natural Science Foundation of China under Grant 61701245, in part by The Startup Foundation for Introducing Talent of NUIST 2243141701030, in part by A Project Funded by the Priority Academic Program Development of Jiangsu Higher Education Institutions.

\section*{References}
\small{
\bibliographystyle{IEEEtran}

}
\end{document}